\pdfoutput=1
\documentclass{article}
\PassOptionsToPackage{numbers, compress}{natbib}

\usepackage[preprint]{neurips_2026}

\usepackage[utf8]{inputenc}
\usepackage[T1]{fontenc}
\usepackage{hyperref}
\usepackage{url}
\usepackage{booktabs}
\usepackage{amsfonts}
\usepackage{amsmath}
\usepackage{amsthm}
\usepackage{nicefrac}
\usepackage{microtype}
\usepackage{xcolor}
\usepackage{graphicx}
\usepackage{algorithm}
\usepackage{algorithmic}
\usepackage{multirow}
\usepackage{makecell}
\usepackage{colortbl}
\usepackage{enumitem}
\usepackage{float}
\usepackage{wrapfig}
\usepackage{caption}
\usepackage[most]{tcolorbox}

\newtheorem{definition}{Definition}

\definecolor{oursHL}{HTML}{E8F5E9}
\definecolor{headerBG}{HTML}{E3F2FD}
\definecolor{negdelta}{HTML}{C62828}
\definecolor{posdelta}{HTML}{2E7D32}
\definecolor{stripeBG}{HTML}{FAFAFA}
\definecolor{leaderHL}{HTML}{FFF8E1}
\title{Not All Skills Help: Measuring and Repairing Agent Knowledge}

\author{%
Yixuan Wang\textsuperscript{1*} \quad
Yiyang Zhou\textsuperscript{1*} \quad
Yiming Liang\textsuperscript{2} \quad
Congyu Zhang\textsuperscript{1} \quad
Fuxiao Liu\textsuperscript{3} \\
\textbf{Jiawei Zhou}\textsuperscript{1} \quad
\textbf{Huaxiu Yao}\textsuperscript{1} \\[6pt]
{\normalfont \textsuperscript{1}UNC Chapel Hill \quad
\textsuperscript{2}Purdue \quad
\textsuperscript{3}NVIDIA}
}

\begin{document}

\maketitle
\let\thefootnote\relax\footnotetext{$^*$Equal contribution.}
\begin{abstract}
LLM agents can improve without weight updates by accumulating natural-language skills from experience, but current systems entrust every decision about which skills to keep and how to apply them to LLM judgment alone. We argue that this conflates two distinct roles: generating a skill from experience is a creative act that judgment handles well, while deciding whether that skill actually helps requires empirical evidence across many tasks. Measuring per-skill causal contributions via randomized masking, we find that skill libraries exhibit pervasive causal heterogeneity: individual skills routinely help on some task types while hurting on others, yet their opposing effects cancel in aggregate, making them invisible to global curation methods. We propose \textsc{\textbf{Assay}}, a framework that separates generation from curation: it computes a per-skill causal attribution on a small development set, restructures the library offline, and suppresses skills with negative predicted effect for each test task. Across seven base models spanning four providers and two benchmarks (AppWorld and $\tau$-bench), \textsc{Assay} consistently improves over prior skill-curation approaches. On AppWorld's hardest split, DeepSeek-V3 achieves 69.3\% task-goal completion (47.4\% relative improvement), a new state of the art among all published methods including weight-tuned approaches. On $\tau$-bench retail, GPT-4.1 improves by 8.7\% relative, advancing past o4-mini, o1, and GPT-4.5 on the public leaderboard without any weight modification. Ablation traces the dominant gain to per-task masking, confirming that the bottleneck is matching skills to tasks at inference time, not removing bad skills globally.
Code is available at \url{https://github.com/aiming-lab/assay}.
\end{abstract}

\section{Introduction}
\label{sec:intro}

The past two years have seen rapid progress in LLM agents that improve without weight updates.
The recipe is simple: let the agent attempt tasks, distill successful trajectories into natural-language \emph{skills} (short rules, heuristics, procedural templates), and inject them into the context window for future tasks~\citep{shinn2023reflexion, zhao2024expel, zhang2025agentic, gupta2025leveraging}.
On interactive benchmarks such as AppWorld~\citep{trivedi2024appworld}, skill-based methods have produced double-digit gains, rivaling methods that fine-tune model weights.
An assumption runs quietly through this work: LLM judgment is a sufficient supervisory signal for the entire skill lifecycle. Generation, retention, and retrieval are all delegated to the same LLM, and nothing ever checks whether a retained skill actually helps.

\begin{figure}[t]
\centering
\includegraphics[width=0.9\textwidth, height=9cm, keepaspectratio]{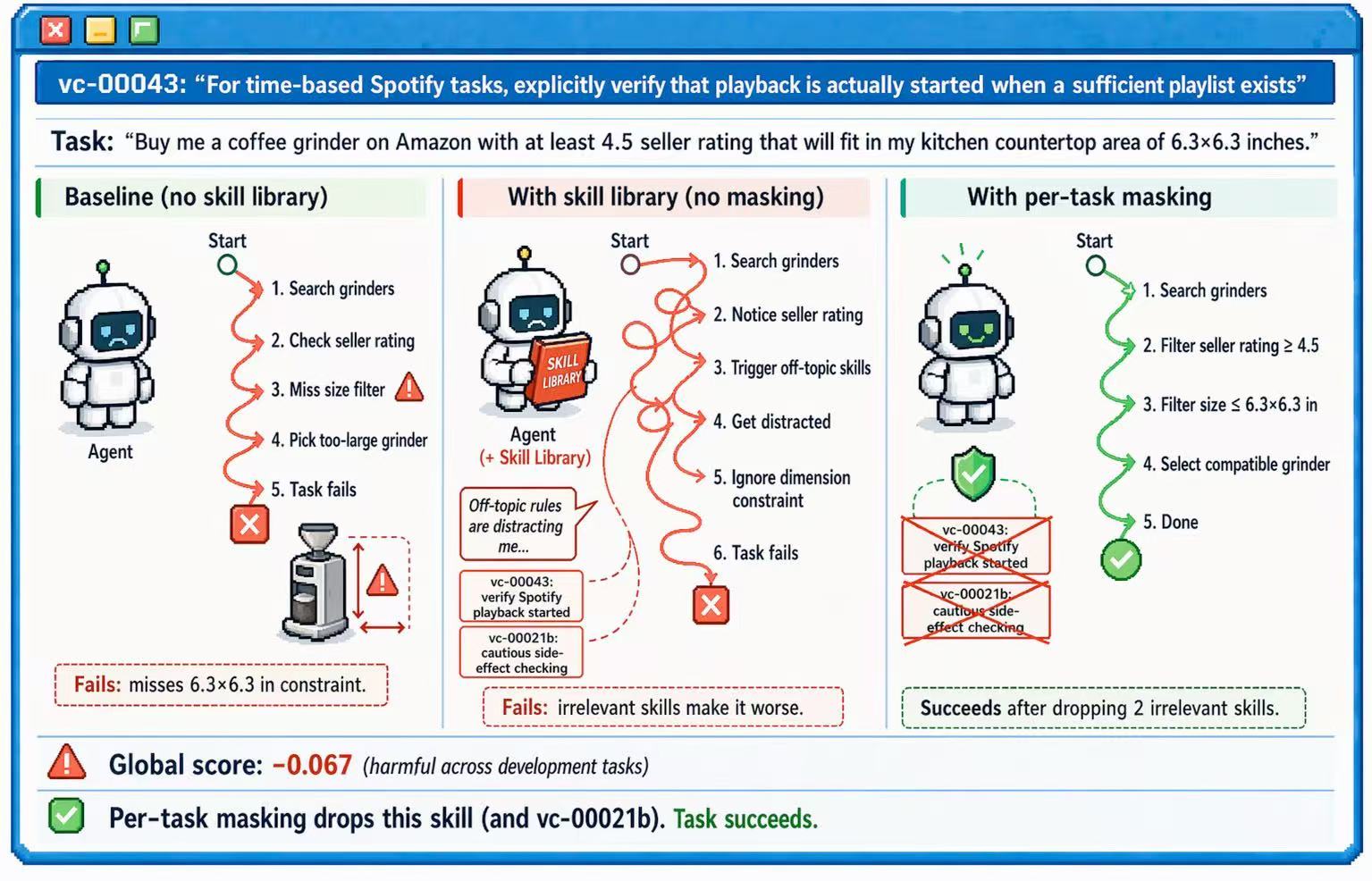}
\caption{\textbf{Skills designed for one domain can harm tasks in another; per-task masking fixes this.}
The task is to purchase a coffee grinder on Amazon that fits a 6.3$\times$6.3-inch countertop and has a seller rating $\geq$4.5.
\emph{Left}: without any skill library, the agent fails to enforce the size constraint.
\emph{Centre}: the full skill library makes things worse: off-topic rules, including a Spotify playback verification rule (\texttt{vc-00043}) and a side-effect checking rule (\texttt{vc-00021b}), distract the agent from the dimensional requirement.
\emph{Right}: per-task masking suppresses these two skills and the agent succeeds.
The causal effect of \texttt{vc-00043} averages $-0.067$ across development tasks (bottom), confirming consistent harm.
}
\label{fig:masking_example}
\vspace{-2em}
\end{figure}

We find that this unchecked accumulation has a systematic downside.
Across seven models and two benchmarks, skills essential on the tasks from which they were learned become pure overhead on tasks where they do not apply.
On AppWorld, rules for multi-step purchases exhaust the step budget on simple single-action tasks; on $\tau$-bench retail~\citep{yao2024tau}, rules from complex exchanges derail straightforward cancellations.
Figure~\ref{fig:masking_example} traces an instance end to end: a Spotify playback verification rule distracts the agent from a dimensional constraint in an Amazon purchasing task; per-task masking suppresses it and the agent succeeds.
When we trace failures across hundreds of tasks, we find that a small set of skills accounts for a disproportionate share of regressions, and that the same skill can help on one task type while hurting on another.

To understand these failures, we measure per-skill causal effects via randomized masking~\citep{covert2021explaining} on held-out tasks.
The resulting attribution reveals \emph{causal heterogeneity}: many skills reverse sign across task types, helping on some and hurting on others.
No single-task judge can detect this, because the reversal is visible only when evidence is aggregated across many tasks.
We therefore separate the two roles: judgment generates skills, and measurement curates them.
We present \textsc{\textbf{Assay}}, a framework that operationalises this separation in three stages: measure per-skill causal effects on held-out tasks via randomized masking, restructure the library offline by splitting heterogeneous skills and retiring inert ones, and personalise the library per test instance by suppressing skills with negative predicted impact.

In summary, our primary contribution is \textsc{Assay}, a framework that provides the first empirical characterisation of causal heterogeneity in agent skill libraries and a practical method for resolving it. Across seven models and two benchmarks, \textsc{Assay} consistently improves over prior curation methods, with DeepSeek-V3 achieving a new state of the art on AppWorld (69.3\% TGC, 47.4\% relative improvement) and GPT-4.1 advancing past o4-mini, o1, and GPT-4.5 on $\tau$-bench without weight modification. Ablation traces the dominant gain to per-task masking, confirming that the bottleneck is matching skills to tasks at inference time, not removing bad skills globally.

\section{\textsc{Assay}: Attribution-Based Skill Selection and Assembly}
\label{sec:method}

\begin{figure}[t]
\centering
\includegraphics[width=1\textwidth]{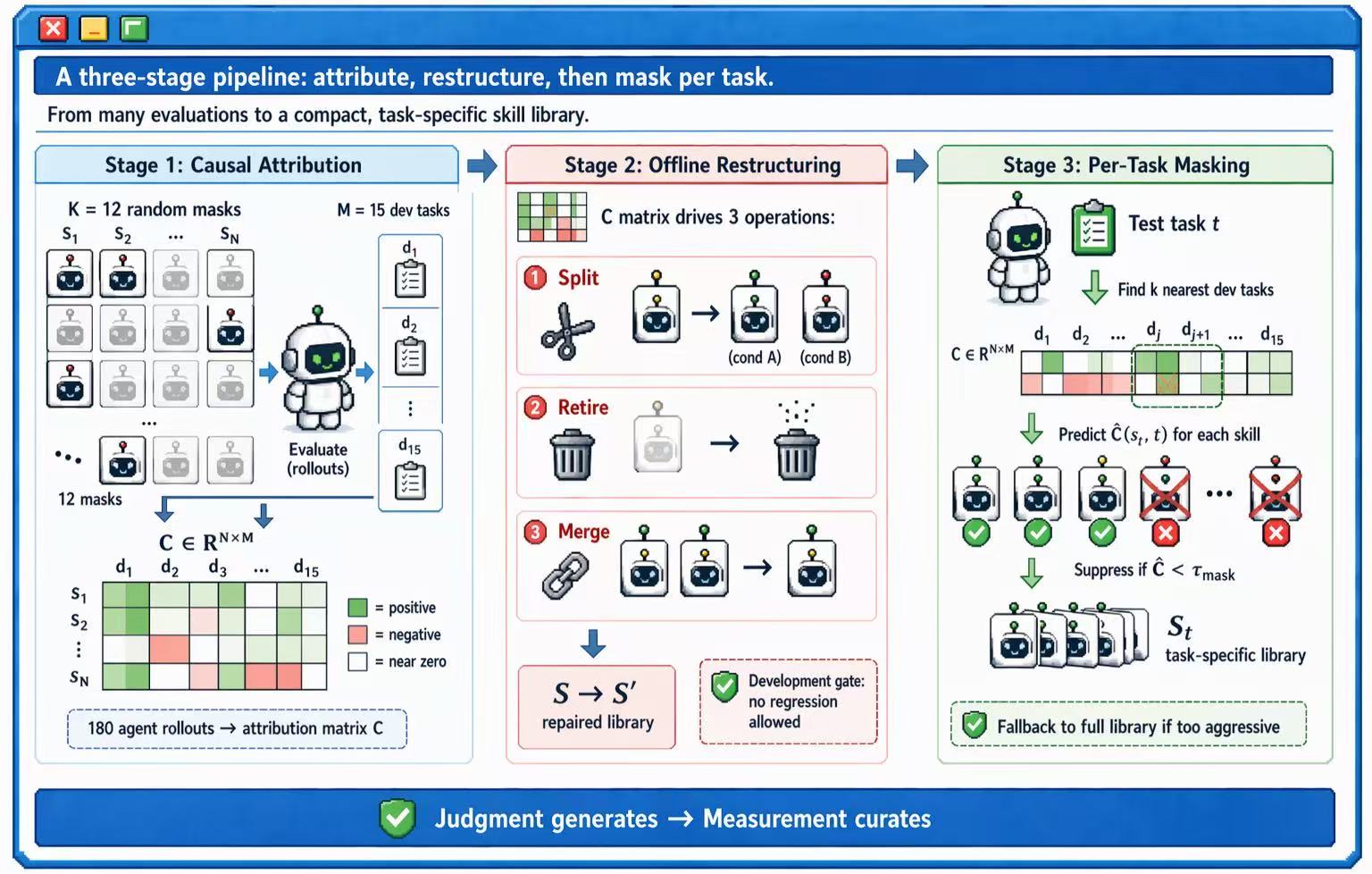}
\caption{\textbf{Framework overview.}
\emph{Stage~1}: randomized masking produces a causal attribution matrix $\mathbf{C} \in \mathbb{R}^{N \times M}$; each cell records whether a skill helps (green) or hurts (red) on a given task.
\emph{Stage~2}: three operations driven by $\mathbf{C}$ (\emph{split}, \emph{retire}, \emph{merge}) restructure the library offline, subject to a development gate.
\emph{Stage~3}: at inference time, per-task masking suppresses skills with negative predicted causal effect, with a fallback to the full library.
}
\label{fig:framework}
\end{figure}

Generating a useful skill from a single task requires creativity; deciding whether that skill actually helps across many tasks requires empirical evidence that no single task can provide.
We operationalise this separation in a framework we call \textsc{\textbf{Assay}}.
The entire framework flows from a single object: a per-skill, per-task causal attribution matrix $\mathbf{C} \in \mathbb{R}^{N \times M}$, computed once on a small held-out set, from which all curation decisions derive.
We describe how $\mathbf{C}$ is measured (\S\ref{sec:attribution}), how it guides offline library restructuring (\S\ref{sec:restructuring}), and how it personalises the library for each test task at inference time (\S\ref{sec:masking}).
Figure~\ref{fig:framework} gives an overview.

\noindent \textbf{Preliminaries.}
We assume access to a skill library $\mathcal{S} = \{s_1, \ldots, s_N\}$ produced by an existing curation pipeline and a held-out development set $\mathcal{D} = \{d_1, \ldots, d_M\}$ disjoint from both training and test splits.
During upstream curation, we apply difficulty-aware ordering: each training task is run twice with the bare agent to estimate difficulty, and tasks are sorted hardest-first so that the curator encounters high-signal failure cases early. The resulting library contains skills automatically distilled from training trajectories.

In addition to these learned skills, we append five hand-written operational templates (prefix \texttt{tpl-}) that capture common procedural patterns such as pagination, data validation, and cross-app identity resolution. These templates are derived from each benchmark's public documentation and training-split failure analysis, and are marked as protected: they are exempt from all subsequent modification and masking (full text in Appendix~\ref{app:templates}).

\subsection{Measuring Causal Effects}
\label{sec:attribution}

A skill library is a collection of natural-language instructions that jointly shape the agent's behaviour.
The challenge in evaluating any single skill is that its effect depends on which other skills are co-present: a verification rule may be harmless alone but harmful alongside a pagination rule that already performs the same check.
To disentangle these interactions, we turn to the simplest tool in causal inference: a randomized experiment.

\noindent \textbf{Randomized masking protocol.}
For each of $K$ independent trials, we construct a random mask $m_k \subseteq \mathcal{S}$ by including each skill independently with probability~$f$ (Bernoulli sampling).
Let $\mathbf{1}[s_j \in m_k]$ denote the inclusion indicator for skill~$s_j$ in mask~$m_k$, and let $o_{k}(d_i) \in \{0,1\}$ denote the binary outcome on development task~$d_i$ when the agent operates under mask~$m_k$.
Define the sets of masks that include and exclude skill~$s_j$ as $\mathcal{M}_j^{+} = \{k : s_j \in m_k\}$ and $\mathcal{M}_j^{-} = \{k : s_j \notin m_k\}$, respectively.
The causal score of skill~$s_j$ on task~$d_i$ is the difference-in-means estimator:
\begin{equation}
\mathbf{C}[j, i] \;=\; \frac{1}{|\mathcal{M}_j^{+}|}\sum_{k \in \mathcal{M}_j^{+}} o_{k}(d_i) \;-\; \frac{1}{|\mathcal{M}_j^{-}|}\sum_{k \in \mathcal{M}_j^{-}} o_{k}(d_i).
\label{eq:causal}
\end{equation}
Under Bernoulli sampling, skills are included independently, so $\mathbf{C}[j,i]$ is an unbiased estimator of the average treatment effect (ATE) of including skill~$s_j$ on task~$d_i$, marginalised over the distribution of co-occurring skills.
The variance of this estimator is bounded by
\begin{equation}
\mathrm{Var}\bigl(\mathbf{C}[j,i]\bigr) \;\leq\; \frac{1}{4}\!\left(\frac{1}{|\mathcal{M}_j^{+}|} + \frac{1}{|\mathcal{M}_j^{-}|}\right),
\label{eq:variance}
\end{equation}
since $o_k(d_i) \in \{0,1\}$ implies $\mathrm{Var}(o_k(d_i)) \leq \tfrac{1}{4}$.
Per-cell variance decreases with more masks; per-task masking further reduces effective noise by averaging over nearest development tasks (Eq.~\ref{eq:predicted}).
All skills participate in attribution without exception; prefix-based protection applies only at the downstream masking stage (\S\ref{sec:masking}).
Specific hyperparameter choices and statistical validation are reported in \S\ref{sec:setup}.

\noindent \textbf{Row statistics.}
Given the full matrix $\mathbf{C}$, we derive two summary statistics from each row that drive all subsequent curation decisions.
The \emph{global causal score} of skill~$s_j$ is the row mean:
\begin{equation}
\bar{C}(j) \;=\; \frac{1}{M}\sum_{i=1}^{M} \mathbf{C}[j, i],
\label{eq:global}
\end{equation}
capturing the average marginal contribution of~$s_j$ across all development tasks.
The \emph{causal heterogeneity} of skill~$s_j$ is the row range:
\begin{equation}
H(s_j) \;=\; \max_{i}\, \mathbf{C}[j,i] \;-\; \min_{i}\, \mathbf{C}[j,i],
\label{eq:heterogeneity}
\end{equation}
measuring the degree to which the skill's effect varies across tasks.

\begin{definition}[Causally heterogeneous skill]
\label{def:heterogeneous}
A skill~$s_j$ is \emph{causally heterogeneous} at threshold~$\tau$ if $H(s_j) \geq \tau$, i.e., its causal effect reverses or substantially varies across development tasks.
\end{definition}

A skill with high $H(s_j)$ but near-zero $\bar{C}(j)$ is the most dangerous: it helps on some tasks and hurts on others, but its positive and negative effects cancel in aggregate, making it invisible to any curation method that evaluates skills by their global score alone.
The full attribution is computed once per base model, because the causal structure of a skill library depends on the model that interprets it.

\subsection{Offline Library Restructuring}
\label{sec:restructuring}

The attribution matrix $\mathbf{C}$ reveals which skills are problematic, but measurement alone does not fix the library. A heterogeneous skill that helps on some tasks and hurts on others cannot simply be removed without losing its beneficial effects. Instead, the library must be restructured so that each skill's applicability conditions are made explicit.
The row statistics $\bar{C}(j)$ and $H(s_j)$ partition skills into three regimes: uniformly beneficial ($\bar{C}(j)$ positive, $H(s_j)$ small), negligible ($|\bar{C}(j)|$ small, $H(s_j)$ small), and causally heterogeneous ($H(s_j) \geq \tau_{\text{split}}$).
We apply three operations targeting each regime in turn: \emph{split} resolves heterogeneous skills into conditional variants, \emph{retire} removes negligible ones, and \emph{merge} deduplicates near-identical skills introduced by splitting. The order is chosen to prevent information loss.

\noindent \textbf{Step 1: Split.}
For each skill~$s_j$ with $H(s_j) \geq \tau_{\text{split}}$, we use the base LLM to rewrite $s_j$ into two conditional variants, each with an explicit trigger condition specifying when it should apply. The rewriting is guided by the causal score vector $(\mathbf{C}[j,1], \ldots, \mathbf{C}[j,M])$: one variant targets tasks where the skill helps, and the other targets tasks where it hurts.
Each rewritten pair must pass a \emph{development gate}: the restructured library must achieve pass rate $\geq$ that of the original on all $M$ attribution tasks.
If it does not, the original skill is kept.
Splitting runs first because a causally heterogeneous skill has $\bar{C}(j) \approx 0$ (positive and negative effects cancel) and would be incorrectly retired if retirement ran first.
At most $\tau_{\text{max\_split}}$ candidates are processed; this is the one point where LLM judgment re-enters curation, bounded in scope and empirically validated.

\noindent \textbf{Step 2: Retire.}
Once heterogeneous skills have been resolved, retirement targets the remaining low-signal skills.
Any skill with $|\bar{C}(j)| < \tau_{\text{retire}}$ is removed.

\noindent \textbf{Step 3: Merge.}
Splitting may introduce near-duplicate variants.
Remaining skills are embedded; pairs exceeding cosine similarity $\tau_{\text{merge}}$ are clustered, and the highest-scoring member of each cluster (by $\bar{C}$) is retained.
Merging runs last precisely because it must operate on the library that splitting and retirement have already shaped.

\subsection{Per-Task Causal Masking}
\label{sec:masking}

Offline restructuring produces a single repaired library $\mathcal{S}'$, but a static library cannot account for the full diversity of test conditions.
The core limitation of existing skill-application methods, whether they inject all skills or retrieve by semantic similarity, is that they cannot distinguish a \emph{helpful} skill from a \emph{harmful} one: the two look identical in embedding space.
We address this by framing skill selection at inference time as a per-task risk minimization problem: for each test task, predict which skills would help and which would hurt, then suppress the harmful ones.

\noindent \textbf{Predicted causal effect.}
The key idea is to transfer causal evidence from similar development tasks to the new test task. Given a test task~$t$ with instruction embedding $\mathbf{e}_t \in \mathbb{R}^d$, we identify the $k$ nearest development tasks $\mathcal{N}(t) \subset \mathcal{D}$ by cosine similarity and compute attention weights via a temperature-scaled softmax:
\begin{equation}
w_i(t) \;=\; \frac{\exp\bigl(\tau \cdot \cos(\mathbf{e}_t,\, \mathbf{e}_{d_i})\bigr)}{\displaystyle\sum_{i' \in \mathcal{N}(t)} \exp\bigl(\tau \cdot \cos(\mathbf{e}_t,\, \mathbf{e}_{d_{i'}})\bigr)},
\qquad i \in \mathcal{N}(t),
\label{eq:weights}
\end{equation}
where $\tau$ is a temperature parameter that concentrates weight on the closest neighbours.
The predicted causal effect of skill~$s_j$ on task~$t$ is a kernel-weighted projection of the $j$-th row of $\mathbf{C}$ onto the task:
\begin{equation}
\hat{C}(s_j,\, t) \;=\; \sum_{i \in \mathcal{N}(t)} w_i(t) \cdot \mathbf{C}[j, i] \;=\; \mathbf{C}[j, :]\, \mathbf{w}(t),
\label{eq:predicted}
\end{equation}
where $\mathbf{w}(t) \in \mathbb{R}^M$ is the weight vector (zero outside $\mathcal{N}(t)$).
This can be read as a soft retrieval over the attribution matrix: each test task induces a different linear combination of the development-task columns of $\mathbf{C}$, producing a task-specific causal profile for every skill.

\noindent \textbf{Risk-minimizing masking rule.}
Given the predicted causal effect for each skill, we suppress skills that are predicted to hurt while retaining all potentially helpful ones.
Under the approximation that skills contribute independently to task success, the expected harm of including skill~$s_j$ for task~$t$ is proportional to $\max(0, -\hat{C}(s_j, t))$.
A removal-only design minimises this expected harm:
\begin{equation}
\mathcal{S}_t \;=\; \bigl\{s_j \in \mathcal{S}' : \hat{C}(s_j, t) \geq \tau_{\text{mask}} \;\;\text{or}\;\; s_j \in \mathcal{S}_{\text{protected}}\bigr\},
\label{eq:masking_rule}
\end{equation}
where $\mathcal{S}_{\text{protected}}$ denotes skills with protected prefixes (\texttt{tpl-}, \texttt{shr-}, \texttt{api-}).
If the filtered set is too small, the full library $\mathcal{S}'$ is used instead, ensuring graceful degradation.
The asymmetry of Eq.~\eqref{eq:masking_rule} is deliberate: missing a critical skill (e.g., a pagination template) causes catastrophic failure, while retaining a mildly harmful skill among many has a diluted effect.
Algorithm~\ref{alg:masking} summarises the procedure.
\begin{algorithm}[t]
\caption{\textsc{Per-Task Causal Masking} (inference time)}
\label{alg:masking}
\begin{algorithmic}[1]
\REQUIRE Test task $t$, restructured library $\mathcal{S}'$, attribution matrix $\mathbf{C} \in \mathbb{R}^{N \times M}$, \\
\hspace{2.2em} development embeddings $\{\mathbf{e}_{d_i}\}_{i=1}^{M}$, parameters $k$, $\tau$, $\tau_{\text{mask}}$, $\tau_{\text{min}}$
\ENSURE Task-specific skill library $\mathcal{S}_t \subseteq \mathcal{S}'$
\vspace{4pt}
\STATE \textbf{// Stage A: Predict per-skill causal effect}
\STATE Compute embedding $\mathbf{e}_t$ of task instruction
\STATE $\mathcal{N}(t) \leftarrow k$ nearest development tasks by $\cos(\mathbf{e}_t, \mathbf{e}_{d_i})$
\STATE $\mathbf{w}(t) \leftarrow \text{softmax}\bigl(\tau \cdot [\cos(\mathbf{e}_t, \mathbf{e}_{d_i})]_{i \in \mathcal{N}(t)}\bigr)$ \hfill $\triangleright$ Eq.~\eqref{eq:weights}
\FOR{each skill $s_j \in \mathcal{S}'$}
  \STATE $\hat{C}(s_j, t) \leftarrow \mathbf{C}[j,:]\,\mathbf{w}(t)$ \hfill $\triangleright$ Eq.~\eqref{eq:predicted}
\ENDFOR
\vspace{4pt}
\STATE \textbf{// Stage B: Risk-minimizing filtering}
\STATE $\mathcal{S}_t \leftarrow \{s_j \in \mathcal{S}' : \hat{C}(s_j, t) \geq \tau_{\text{mask}} \;\text{or}\; s_j \in \mathcal{S}_{\text{protected}}\}$ \hfill $\triangleright$ Eq.~\eqref{eq:masking_rule}
\vspace{4pt}
\STATE \textbf{// Stage C: Graceful fallback}
\IF{$|\mathcal{S}_t| < \tau_{\text{min}}$}
  \STATE $\mathcal{S}_t \leftarrow \mathcal{S}'$ \hfill $\triangleright$ Revert to full library
\ENDIF
\RETURN $\mathcal{S}_t$
\end{algorithmic}
\end{algorithm}
\noindent \textbf{Summary.}
The three stages form a coherent pipeline unified by a single principle: judgment generates candidate skills, and measurement curates them.
Every stage includes a structural fallback (offline restructuring rolls back failing splits; per-task masking reverts to the full library when too aggressive), so the pipeline cannot degrade performance below any prefix of stages.
Because the framework operates entirely at inference time and requires only a skill library and a small development set, it can be applied on top of any existing skill-generation method.

\section{Experiments}
\label{sec:setup}

We evaluate on two benchmarks, seven base models spanning four providers, and two agent architectures, applying the same pipeline and hyperparameters throughout. Our experiments address the following questions: (1)~Does measurement-driven curation generalise across models and benchmarks? (2)~Where do the gains concentrate, and where does uncurated skill injection cause harm? (3)~Which stages of the pipeline matter most?

\subsection{Experimental Setup}

\noindent \textbf{Benchmarks.}
\textbf{AppWorld}~\citep{trivedi2024appworld} simulates nine consumer applications (email, calendar, Venmo, Spotify, Amazon, etc.) in which the agent composes multi-step Python API calls via a code REPL.
We evaluate on two official test splits: \textit{test\_normal} (168~tasks) and \textit{test\_challenge} (417~tasks), both spanning difficulty levels~1--3.
\textit{test\_challenge} has a higher concentration of level-3 tasks (47\% vs.\ 37\%), making it substantially harder on average.
The primary metric is Task Goal Completion (TGC); we also report Sub-Goal Completion (SGC; Appendix~\ref{app:sgc}).
The agent is a ReAct agent~\citep{yao2022react} with a 40-step budget.
\textbf{$\tau$-bench}~\citep{yao2024tau} simulates retail customer service through function-calling tools.
The agent converses with an LLM-simulated customer (GPT-4o, temperature~0) and must satisfy requests while adhering to a policy document.
We evaluate on the retail domain (115~tasks) with exact database-state match; partial credit is not awarded.
The agent is the benchmark's native \texttt{ToolCallingAgent} with a 30-turn budget.
In both benchmarks, the skill library is injected as part of the system prompt and all agent calls use temperature~0.

\noindent \textbf{Models and data.}
We evaluate seven models in standard (non-reasoning) mode: GPT-5.4, GPT-5.1, GPT-4.1, and GPT-4o (OpenAI); DeepSeek-V3 (DeepSeek, open-weight); Claude Sonnet~4.5 (Anthropic); and Gemini 2.5 Pro (Google). We deliberately exclude reasoning variants to isolate the effect of skill curation from chain-of-thought reasoning.
Each model runs the complete pipeline independently; the attribution matrix $\mathbf{C}$ is recomputed per model because the causal structure of a skill library depends on the model that interprets it (\S\ref{sec:attribution}).
Data is partitioned into strictly disjoint sets: AppWorld uses 90~training, 15~development (of 57), and $168 {+} 417$ test tasks; $\tau$-bench uses 500~training, 15~development (of 20), and 115~test.

\noindent \textbf{Baselines.}
For each model on AppWorld, we compare against the best available published method:
ACE~\citep{zhang2025agentic} for GPT-5.1 and DeepSeek-V3,
CUGA~\citep{marreed2025towards} for GPT-4.1, and
\citet{gupta2025leveraging} for GPT-4o.
We additionally report the current AppWorld leaderboard leader.\footnote{Alibaba Cloud ApsaraLab, AppWorld leaderboard submission (February 2026, Qwen3-14B, weight-tuned). No accompanying publication is available as of this writing.}
On $\tau$-bench, we compare against each model's unaugmented baseline, since no prior skill-based method has reported results on this benchmark under the same evaluation protocol.
For GPT-5.4, Claude Sonnet~4.5, and Gemini 2.5 Pro, no prior skill-based method has reported results on AppWorld either; we compare against the unaugmented ReAct baseline for these models.

\noindent \textbf{Attribution parameters.}
The attribution uses $K{=}12$ random masks with inclusion probability $f{=}0.4$, yielding approximately 5 masks per skill and a per-cell standard deviation of at most 0.29.
Per-task masking averages over $k{=}8$ nearest development tasks, reducing effective per-decision noise to $\sigma \approx 0.10$.
Bootstrap analysis confirms that 97.2\% of masking decisions are directionally stable under mask resampling (Appendix~\ref{app:stat_validation}).
The full attribution requires 180 agent rollouts per model ($K{=}12$ masks $\times$ $M{=}15$ development tasks).
All remaining hyperparameters are listed in Appendix Table~\ref{tab:hyperparams} (Appendix~\ref{app:reproducibility}); no per-cell tuning is performed.

\subsection{Main Results}
\label{sec:results}


\begin{table}[t]
\caption{\textbf{AppWorld results} (TGC, \%).
$\Delta$ is computed relative to bare ReAct.
\textbf{Bold}: best prompt-based method per column.
{\color{negdelta}Red}: degrades over ReAct.
SGC is reported in Appendix~\ref{app:sgc}.}
\label{tab:main}
\centering
\footnotesize
\setlength{\tabcolsep}{5pt}
\renewcommand{\arraystretch}{1.0}
\begin{tabular}{@{}ll cc cc@{}}
\toprule
& & \multicolumn{2}{c}{\textit{test\_normal} (168)} & \multicolumn{2}{c}{\textit{test\_challenge} (417)} \\
\cmidrule(lr){3-4} \cmidrule(lr){5-6}
\textbf{Model} & \textbf{Method} & TGC & $\Delta$ & TGC & $\Delta$ \\
\midrule
\multirow{3}{*}{GPT-5.1}
 & ReAct               & 61.9 &       & 52.5 &  \\
 & ACE                 & 67.3 & {\color{posdelta}\small$+$5.4}  & {\color{negdelta}49.9} & {\color{negdelta}\small$-$2.6} \\
 \rowcolor{oursHL}
 & \textbf{Ours}       & \textbf{77.4} & {\color{posdelta}\small$+$15.5} & \textbf{66.4} & {\color{posdelta}\small$+$13.9} \\
\addlinespace[1.5pt]
\multirow{3}{*}{DeepSeek-V3}
 & ReAct               & 69.1 &       & 47.0 &  \\
 & ACE                 & 78.0 & {\color{posdelta}\small$+$8.9}  & 63.1 & {\color{posdelta}\small$+$16.1} \\
 \rowcolor{oursHL}
 & \textbf{Ours}       & \textbf{83.3} & {\color{posdelta}\small$+$14.2} & \textbf{69.3} & {\color{posdelta}\small$+$22.3} \\
\addlinespace[1.5pt]
\multirow{3}{*}{GPT-4.1}
 & ReAct               & 66.7 &       & 50.4 &  \\
 & CUGA                & 73.2 & {\color{posdelta}\small$+$6.5}  & 57.6 & {\color{posdelta}\small$+$7.2} \\
 \rowcolor{oursHL}
 & \textbf{Ours}       & \textbf{75.6} & {\color{posdelta}\small$+$8.9}  & \textbf{64.0} & {\color{posdelta}\small$+$13.6} \\
\addlinespace[1.5pt]
\multirow{3}{*}{GPT-4o}
 & ReAct               & 48.8 &       & 30.2 &  \\
 & \citeauthor{gupta2025leveraging} & 68.5 & {\color{posdelta}\small$+$19.7} & 38.9 & {\color{posdelta}\small$+$8.7} \\
 \rowcolor{oursHL}
 & \textbf{Ours}       & \textbf{71.4} & {\color{posdelta}\small$+$22.6} & \textbf{41.0} & {\color{posdelta}\small$+$10.8} \\
\addlinespace[1.5pt]
GPT-5.4
 & ReAct               & 82.1 &       & 81.1 &  \\
 \rowcolor{oursHL}
 & \textbf{Ours}       & \textbf{88.7} & {\color{posdelta}\small$+$6.6}  & \textbf{85.4} & {\color{posdelta}\small$+$4.3} \\
\addlinespace[1.5pt]
Sonnet 4.5
 & ReAct               & 83.9 &       & 70.3 &  \\
 \rowcolor{oursHL}
 & \textbf{Ours}       & \textbf{89.3} & {\color{posdelta}\small$+$5.4}  & \textbf{75.3} & {\color{posdelta}\small$+$5.0} \\
\addlinespace[1.5pt]
Gemini 2.5
 & ReAct               & 72.6 &       & 49.4 &  \\
 \rowcolor{oursHL}
 & \textbf{Ours}       & \textbf{81.0} & {\color{posdelta}\small$+$8.4}  & \textbf{54.9} & {\color{posdelta}\small$+$5.5} \\
\midrule
\rowcolor{leaderHL}
\multicolumn{2}{@{}l}{\small\textit{Leaderboard best$^*$ (Qwen3-14B, wt-tuned)}}
 & \textit{86.9} & & \textit{67.6} & \\
\bottomrule
\multicolumn{6}{@{}l}{\footnotesize $^*$No accompanying publication; see footnote in \S\ref{sec:setup}.}
\end{tabular}
\vspace{-2em}
\end{table}

\noindent \textbf{AppWorld.}
Table~\ref{tab:main} presents results on both AppWorld splits. Every model improves over its respective baseline on both splits.
On \textit{test\_normal}, GPT-5.1 gains 10.1~points over ACE (67.3$\to$77.4) and DeepSeek-V3 gains 5.3~points (78.0$\to$83.3), both from the same upstream library with only curation changed. GPT-4.1 and GPT-4o improve over CUGA and \citeauthor{gupta2025leveraging} respectively, baselines that already incorporate structured retrieval or demonstration selection. Three additional models (GPT-5.4, Sonnet~4.5, Gemini~2.5 Pro) confirm generality across four providers with gains ranging from $+$4.3 to $+$8.4 across both splits, with no per-provider tuning.
On the harder split, DeepSeek-V3 achieves 69.3\%~TGC, a new state of the art among all published methods including weight-tuned approaches, representing a 47.4\% relative improvement over bare ReAct.

\begin{wrapfigure}{r}{0.58\textwidth}
\vspace{-12pt}
\centering
\includegraphics[width=0.58\textwidth]{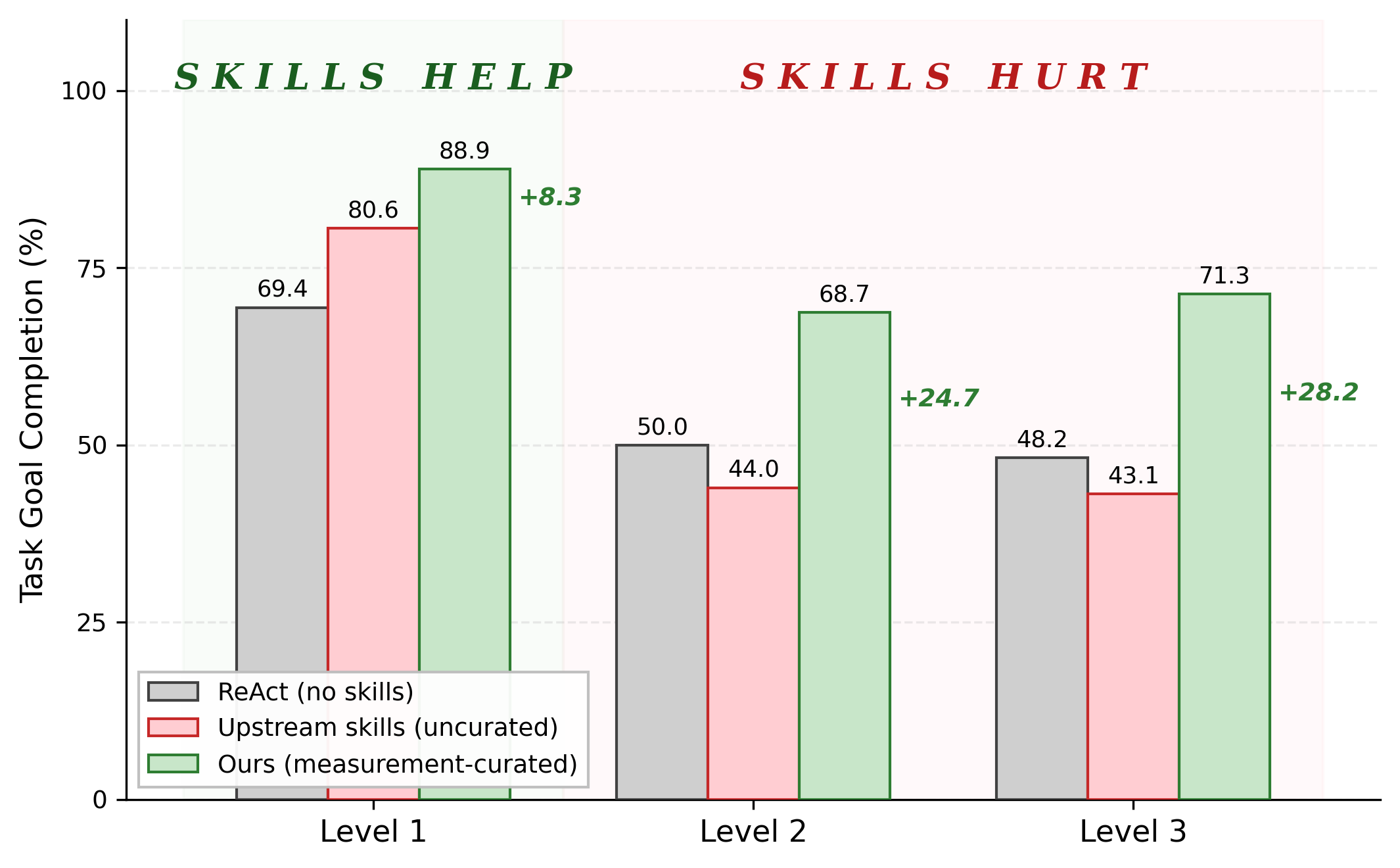}
\captionsetup{font=small}
\caption{\textbf{Per-difficulty breakdown on AppWorld \textit{test\_challenge}} (GPT-5.1, 417 tasks).
The uncurated skill library (red) improves Level~1 but degrades Level~2 and~3.
Our method (green) recovers at every level.}
\label{fig:difficulty}
\vspace{-8pt}
\end{wrapfigure}
The most revealing result is not the gain but the regression that precedes it. On \textit{test\_challenge}, the upstream skill library \emph{decreases} GPT-5.1's TGC from 52.5\% to 49.9\%: a library designed to help has made the agent strictly worse. Our method reverses this regression and improves by 26.5\% relative beyond bare ReAct. Figure~\ref{fig:difficulty} reveals where the harm concentrates: on Level~2 and Level~3 tasks, the uncurated library degrades performance, while Level~1 tasks already benefit. Per-task masking recovers the degraded levels, with the largest gain on Level~3 (43.1$\to$71.3, a 65.4\% relative improvement), confirming that the method's value is greatest where uncurated libraries do the most damage. A reverse-masking control (suppressing the most \emph{positively}-scoring skills instead) degrades performance, confirming that the direction of masking, not mere context reduction, drives the improvement.

\noindent \textbf{$\tau$-bench.}
Table~\ref{tab:combined} evaluates whether the framework transfers to a qualitatively different setting: conversational rather than code-based, function-calling rather than REPL, and with a simulated human in the loop.

\begin{wraptable}{H}{0.52\textwidth}
\vspace{-12pt}
\centering
\small
\setlength{\tabcolsep}{5pt}
\renewcommand{\arraystretch}{1.12}
\begin{tabular}{@{}r l c@{}}
\toprule
\textbf{Rank} & \textbf{Method} & \textbf{Score} \\
\midrule
1--5 & Claude family & 80.5--86.2 \\
\rowcolor{oursHL}
     & \textbf{Ours (GPT-5.4)} {\footnotesize\color{posdelta}$\uparrow$7.0} & \textbf{80.9} \\
6    & GLM-4.5       & 79.7 \\
7    & GLM-4.5-Air   & 77.9 \\
8    & Qwen3-Coder 480B & 77.5 \\
\rowcolor{oursHL}
     & \textbf{Ours (GPT-4.1)} {\footnotesize\color{posdelta}$\uparrow$5.9} & \textbf{73.9} \\
\rowcolor{oursHL}
     & \textbf{Ours (Gemini 2.5)} {\footnotesize\color{posdelta}$\uparrow$8.7} & \textbf{73.9} \\
\rowcolor{stripeBG}
9    & o4-mini       & 71.8 \\
10   & o1            & 70.8 \\
\rowcolor{stripeBG}
13   & GPT-4.5       & 68.4 \\
14   & GPT-4.1 {\footnotesize(raw)} & 68.0 \\
\rowcolor{oursHL}
     & \textbf{Ours (DS-V3)} {\footnotesize\color{posdelta}$\uparrow$2.2} & \textbf{66.1} \\
\rowcolor{oursHL}
     & \textbf{Ours (GPT-4o)} {\footnotesize\color{posdelta}$\uparrow$2.2} & \textbf{62.6} \\
     & GPT-4o {\footnotesize(raw)}  & 60.3 \\
\bottomrule
\end{tabular}
\captionsetup{font=small}
\caption{\textbf{$\tau$-bench retail positioning on the public leaderboard.}
Green rows with $\uparrow$ show our method's gain over the raw baseline.
GPT-4.1 advances from rank\,14 to rank\,8--9; GPT-5.4 reaches the top-5 range.
Two models (GPT-5.1, Sonnet~4.5) show zero gain (\S\ref{app:limitations}).}
\label{tab:combined}
\vspace{-8pt}
\end{wraptable}

The framework transfers successfully. GPT-4.1 improves by 8.7\% relative (68.0\%$\to$73.9\%), advancing from rank~14 to the rank~8--9 range on the leaderboard, past o4-mini, o1, and GPT-4.5, without any weight modification. GPT-5.4 gains 9.5\% relative (73.9\%$\to$80.9\%) and Gemini~2.5 Pro gains 13.3\% relative (65.2\%$\to$73.9\%).
Two models show zero gain: GPT-5.1 (62.6\%$\to$62.6\%) and Sonnet~4.5 (73.0\%$\to$73.0\%); we analyse these boundary conditions in Appendix~\ref{app:limitations} and attribute them to high baseline competence that saturates the benefit of prompt-time skill injection.
Gains concentrate on multi-step \textit{return} and \textit{cancel} tasks (Appendix Table~\ref{tab:taubench_action}), where the skill library encodes procedural knowledge that the unaugmented agent must otherwise rediscover from the policy document on every task.
\subsection{Ablation Study}

Table~\ref{tab:ablation} (Appendix~\ref{app:ablation}) isolates each component's contribution via sequential ablation on GPT-5.1 / AppWorld \textit{test\_normal}.
Templates provide a 9.7\% relative improvement, confirming that domain-agnostic operational scaffolding carries measurable value.
Offline restructuring adds a further 2.9\% relative gain by resolving heterogeneous skills and retiring inert ones.
Per-task masking contributes the largest single increment (10.7\% relative improvement), consistent with the central finding of this paper: the bottleneck is not which skills are in the library, but which skills each task should see.
The full pipeline achieves a 25.0\% relative improvement over bare ReAct.

\subsection{Causal Heterogeneity in Individual Skills}

The aggregate results above show that curation helps; this section examines \emph{why} by tracing causal heterogeneity to individual skills. Two examples from the GPT-5.1 attribution on AppWorld illustrate the phenomenon.

\begin{tcolorbox}[
  colback=headerBG!30,
  colframe=headerBG!80!black,
  sharp corners,
  left=2mm,right=2mm,top=2mm,bottom=2mm,
  title={\small\textbf{Example 1: Contact Validation Rule (\texttt{vc-00054})}}]
\small
\textbf{Skill.} ``Validate that note names map unambiguously to contacts before creating Venmo requests.'' \\[3pt]
\textbf{Global score.} $\bar{C} = -0.03$ (near zero, invisible to global curation). \\[3pt]
{\color{posdelta}\textbf{Helps}} ($+0.50$): shared-expense reconciliation tasks, where the validation catches real name ambiguities. \\[3pt]
{\color{negdelta}\textbf{Hurts}} ($-0.67$): single-app tasks, where it forces the agent to cross-reference contacts irrelevant to the task.
\end{tcolorbox}
\vspace{-4pt}
\begin{tcolorbox}[
  colback=headerBG!30,
  colframe=headerBG!80!black,
  sharp corners,
  left=2mm,right=2mm,top=2mm,bottom=2mm,
  title={\small\textbf{Example 2: Response Checking Rule (\texttt{vc-00021a})}}]
\small
\textbf{Skill.} ``For repeated side-effecting API calls, capture and check each response.'' \\[3pt]
\textbf{Global score.} $\bar{C} = +0.05$ (near zero, invisible to global curation). \\[3pt]
{\color{posdelta}\textbf{Helps}} ($+0.71$): mutation-heavy tasks, where unchecked API calls cause silent failures. \\[3pt]
{\color{negdelta}\textbf{Hurts}} ($-0.23$): read-only tasks, where the checks are pure overhead.
\end{tcolorbox}

\vspace{2pt}
Both skills would survive any global curation threshold. Only per-task measurement reveals their conditional nature, and only per-task masking can suppress them selectively. These examples confirm that causal heterogeneity is concrete and interpretable: skills encode assumptions about the task context, and when those assumptions are violated, the skill becomes harmful. Additional examples are in Appendix~\ref{app:skill_examples}.

\section{Related Work}
\label{sec:related}

\noindent \textbf{Skill generation and curation.}
A growing line of work improves agent performance by accumulating experience into the agent's context rather than its weights.
The methods differ in mechanism: Reflexion~\citep{shinn2023reflexion} maintains verbal self-reflections in an episodic memory buffer, ExpeL~\citep{zhao2024expel} extracts reusable insights by comparing successful and failed trajectories, ACE~\citep{zhang2025agentic} grows evolving playbooks through a modular generate-reflect-curate loop, and CUGA~\citep{marreed2025towards} adopts a hierarchical planner-executor architecture with context enrichment. More recently, SkillNet~\citep{liang2026skillnet} provides end-to-end tooling to create, evaluate, and connect skills within a unified ontology, and CoEvoSkills~\citep{zhang2026coevoskills} co-evolves a skill generator and surrogate verifier without access to ground-truth tests. These methods share a structural property: every decision in the skill lifecycle, from generation to retention to application, is made by LLM judgment operating within individual tasks. No method aggregates evidence across tasks to verify that a retained skill actually helps.
Our work treats the output of any such pipeline as raw material for a second, empirical curation stage.

\noindent \textbf{Skill retrieval and application.}
The same reliance on single-task judgment extends to how skills are surfaced at test time.
Retrieval-augmented generation~\citep{lewis2020retrieval} and its agent-oriented variants~\citep{zhang2024agentohana} select context by embedding similarity, implicitly equating topical relevance with helpfulness.
\citet{gupta2025leveraging} extend BERTScore-Recall to set-level demonstration selection for in-context learning in agentic tasks, and \citet{su2026skillretrieval} study skill retrieval augmentation at scale as agent skill libraries grow to thousands of entries.
Our causal attribution reveals that this equation can be precisely wrong: a skill about ``verifying list contents before iteration'' is semantically close to a single-record lookup yet hurts on such tasks by inducing unnecessary verification steps.
Per-task filtering addresses this by conditioning on predicted causal effect rather than similarity.

\noindent \textbf{Skill optimization beyond judgment.}
Several methods go beyond single-task judgment to optimize skill libraries. Voyager~\citep{wangvoyager} builds an expanding skill library but only adds skills, never removing or conditioning them. SkillRL~\citep{xia2026skillrl} uses reinforcement learning to recursively refine skills through failure analysis, treating the library as a dynamic component co-evolving with the agent policy. EvolveR~\citep{wu2025evolver} closes an experience-driven evolution loop, and Agentic Memory~\citep{yu2026agenticmemory} optimizes memory management with GRPO. SkillClaw~\citep{ma2026skillclaw} enables collective skill evolution by aggregating trajectories across users to identify recurring behavioral patterns, and GraSP~\citep{xia2026grasp} compiles flat skill sets into typed directed acyclic graphs, observing that providing agents with more skills does not monotonically improve performance. Weight-based methods such as SAGE~\citep{wang2025reinforcement} and FireAct~\citep{chen2023fireact} internalise skills into model parameters via RL or trajectory fine-tuning, avoiding context-window limitations but requiring retraining for each base model.
Our approach is complementary to all of the above: it operates at inference time, requires no weight updates, and can be layered on top of any skill-generation pipeline. The key distinction is that we measure per-skill causal effects via randomized masking, following the logic of Shapley-value attribution~\citep{lundberg2017unified} and randomized ablation~\citep{covert2021explaining}, but applied to natural-language skill instructions rather than model components.

\section{Conclusion}
\label{sec:conclusion}

We presented \textsc{Assay}, a framework that separates skill generation from skill curation by measuring per-skill causal effects via randomized masking. The causal attribution reveals pervasive heterogeneity in skill libraries, and per-task masking resolves it: on AppWorld, all seven models improve across four providers, with DeepSeek-V3 achieving a new state of the art (69.3\%, 47.4\% relative improvement); on $\tau$-bench, GPT-4.1 advances past o4-mini, o1, and GPT-4.5 without weight modification. Ablation confirms that the bottleneck is matching skills to tasks at inference time, not removing bad skills globally. Two null results (GPT-5.1, Sonnet~4.5 on $\tau$-bench) highlight a boundary: as base model competence strengthens, prompt-time skill injection yields diminishing returns. Extending the framework to online settings where skills are continuously added is a promising direction for future work.
\bibliographystyle{plainnat}
\bibliography{references}

@article{shinn2023reflexion,
  title={Reflexion: Language agents with verbal reinforcement learning},
  author={Shinn, Noah and Cassano, Federico and Berman, Edward and Gopinath, Ashwin and Narasimhan, Karthik and Yao, Shunyu},
  journal={Advances in neural information processing systems},
  volume={36},
  pages={8634--8652},
  year={2023}
}

@inproceedings{zhao2024expel,
  title={Expel: Llm agents are experiential learners},
  author={Zhao, Andrew and Huang, Daniel and Xu, Quentin and Lin, Matthieu and Liu, Yong-Jin and Huang, Gao},
  booktitle={Proceedings of the AAAI Conference on Artificial Intelligence},
  volume={38},
  number={17},
  pages={19632--19642},
  year={2024}
}

@article{zhang2025agentic,
  title={Agentic context engineering: Evolving contexts for self-improving language models},
  author={Zhang, Qizheng and Hu, Changran and Upasani, Shubhangi and Ma, Boyuan and Hong, Fenglu and Kamanuru, Vamsidhar and Rainton, Jay and Wu, Chen and Ji, Mengmeng and Li, Hanchen and Thakker, Urmish and Zou, James and Olukotun, Kunle},
  journal={arXiv preprint arXiv:2510.04618},
  year={2025}
}

@article{gupta2025leveraging,
  title={Leveraging in-context learning for language model agents},
  author={Gupta, Shivanshu and Singh, Sameer and Sabharwal, Ashish and Khot, Tushar and Bogin, Ben},
  journal={arXiv preprint arXiv:2506.13109},
  year={2025}
}

@article{marreed2025towards,
  title={Towards enterprise-ready computer using generalist agent},
  author={Marreed, Sami and Oved, Alon and Yaeli, Avi and Shlomov, Segev and Levy, Ido and Akrabi, Offer and Sela, Aviad and Adi, Asaf and Mashkif, Nir},
  journal={arXiv preprint arXiv:2503.01861},
  year={2025}
}

@article{wang2025reinforcement,
  title={Reinforcement learning for self-improving agent with skill library},
  author={Wang, Jiongxiao and Yan, Qiaojing and Wang, Yawei and Tian, Yijun and Mishra, Soumya Smruti and Xu, Zhichao and Gandhi, Megha and Xu, Panpan and Cheong, Lin Lee},
  journal={arXiv preprint arXiv:2512.17102},
  year={2025}
}

@article{chen2023fireact,
  title={Fireact: Toward language agent fine-tuning},
  author={Chen, Baian and Shu, Chang and Shareghi, Ehsan and Collier, Nigel and Narasimhan, Karthik and Yao, Shunyu},
  journal={arXiv preprint arXiv:2310.05915},
  year={2023}
}

@inproceedings{trivedi2024appworld,
  title={Appworld: A controllable world of apps and people for benchmarking interactive coding agents},
  author={Trivedi, Harsh and Khot, Tushar and Hartmann, Mareike and Manku, Ruskin and Dong, Vinty and Li, Edward and Gupta, Shashank and Sabharwal, Ashish and Balasubramanian, Niranjan},
  booktitle={Proceedings of the 62nd Annual Meeting of the Association for Computational Linguistics (Volume 1: Long Papers)},
  pages={16022--16076},
  year={2024}
}

@article{yao2024tau,
  title={$\tau$-bench: A Benchmark for Tool-Agent-User Interaction in Real-World Domains},
  author={Yao, Shunyu and Shinn, Noah and Razavi, Pedram and Narasimhan, Karthik},
  journal={arXiv preprint arXiv:2406.12045},
  year={2024}
}

@inproceedings{yao2022react,
  title={React: Synergizing reasoning and acting in language models},
  author={Yao, Shunyu and Zhao, Jeffrey and Yu, Dian and Du, Nan and Shafran, Izhak and Narasimhan, Karthik R and Cao, Yuan},
  booktitle={The eleventh international conference on learning representations},
  year={2022}
}

@article{lewis2020retrieval,
  title={Retrieval-augmented generation for knowledge-intensive nlp tasks},
  author={Lewis, Patrick and Perez, Ethan and Piktus, Aleksandra and Petroni, Fabio and Karpukhin, Vladimir and Goyal, Naman and K{\"u}ttler, Heinrich and Lewis, Mike and Yih, Wen-tau and Rockt{\"a}schel, Tim and others},
  journal={Advances in neural information processing systems},
  volume={33},
  pages={9459--9474},
  year={2020}
}

@article{zhang2024agentohana,
  title={Agentohana: Design unified data and training pipeline for effective agent learning},
  author={Zhang, Jianguo and Lan, Tian and Murthy, Rithesh and Liu, Zhiwei and Yao, Weiran and Zhu, Ming and Tan, Juntao and Hoang, Thai and Liu, Zuxin and Yang, Liangwei and others},
  journal={arXiv preprint arXiv:2402.15506},
  year={2024}
}

@article{lundberg2017unified,
  title={A unified approach to interpreting model predictions},
  author={Lundberg, Scott M and Lee, Su-In},
  journal={Advances in neural information processing systems},
  volume={30},
  year={2017}
}

@article{covert2021explaining,
  title={Explaining by removing: A unified framework for model explanation},
  author={Covert, Ian and Lundberg, Scott and Lee, Su-In},
  journal={Journal of Machine Learning Research},
  volume={22},
  number={209},
  pages={1--90},
  year={2021}
}

@article{xia2026skillrl,
  title={SkillRL: Evolving Agents via Recursive Skill-Augmented Reinforcement Learning},
  author={Xia, Peng and Chen, Jianwen and Wang, Hanyang and Liu, Jiaqi and Zeng, Kaide and Wang, Yu and Han, Siwei and Zhou, Yiyang and Zhao, Xujiang and Chen, Haifeng and Zheng, Zeyu and Xie, Cihang and Yao, Huaxiu},
  journal={arXiv preprint arXiv:2602.08234},
  year={2026}
}

@article{wu2025evolver,
  title={{EvolveR}: Self-Evolving {LLM} Agents through an Experience-Driven Lifecycle},
  author={Wu, Rong and Wang, Xiaoman and Mei, Jianbiao and Cai, Pinlong and Fu, Daocheng and Yang, Cheng and Wen, Licheng and Yang, Xuemeng and Shen, Yufan and Wang, Yuxin and Shi, Botian},
  journal={arXiv preprint arXiv:2510.16079},
  year={2025}
}

@article{wangvoyager,
  title={Voyager: An Open-Ended Embodied Agent with Large Language Models},
  author={Wang, Guanzhi and Xie, Yuqi and Jiang, Yunfan and Mandlekar, Ajay and Xiao, Chaowei and Zhu, Yuke and Fan, Linxi and Anandkumar, Anima},
  journal={Transactions on Machine Learning Research},
  year={2024}
}

@article{yu2026agenticmemory,
  title={Agentic Memory: Learning Unified Long-Term and Short-Term Memory Management for Large Language Model Agents},
  author={Yu, Yi and Yao, Liuyi and Xie, Yuexiang and Tan, Qingquan and Feng, Jiaqi and Li, Yaliang and Wu, Libing},
  journal={arXiv preprint arXiv:2601.01885},
  year={2026}
}

@article{ma2026skillclaw,
  title={SkillClaw: Let Skills Evolve Collectively with Agentic Evolver},
  author={Ma, Ziyu and Yang, Shidong and Ji, Yuxiang and Wang, Xucong and Wang, Yong and Hu, Yiming and Huang, Tongwen and Chu, Xiangxiang},
  journal={arXiv preprint arXiv:2604.08377},
  year={2026}
}

@article{xia2026grasp,
  title={{GraSP}: Graph-Structured Skill Compositions for {LLM} Agents},
  author={Xia, Tianle and Hu, Lingxiang and Sun, Yiding and Xu, Ming and Xu, Lan and Wang, Siying and Xu, Wei and Jiang, Jie},
  journal={arXiv preprint arXiv:2604.17870},
  year={2026}
}

@article{liang2026skillnet,
  title={SkillNet: Create, Evaluate, and Connect {AI} Skills},
  author={Liang, Yuan and Zhong, Ruobin and Xu, Haoming and Jiang, Chen and Zhong, Yi and Fang, Runnan and Gu, Jia-Chen and Deng, Shumin and Yao, Yunzhi and Wang, Mengru and others},
  journal={arXiv preprint arXiv:2603.04448},
  year={2026}
}

@article{zhang2026coevoskills,
  title={CoEvoSkills: Self-Evolving Agent Skills via Co-Evolutionary Verification},
  author={Zhang, Hanrong and Fan, Shicheng and Zou, Henry Peng and Chen, Yankai and Wang, Zhenting and Zhou, Jiayu and Li, Chengze and Huang, Wei-Chieh and Yao, Yifei and Zheng, Kening and Liu, Xue and Li, Xiaoxiao and Yu, Philip S.},
  journal={arXiv preprint arXiv:2604.01687},
  year={2026}
}

@article{su2026skillretrieval,
  title={Skill Retrieval Augmentation for Agentic {AI}},
  author={Su, Weihang and Long, Jianming and Ai, Qingyao and Tang, Yichen and Wang, Changyue and Tu, Yiteng and Liu, Yiqun},
  journal={arXiv preprint arXiv:2604.24594},
  year={2026}
}

\appendix

\section{Extended Analysis and Discussion}
\label{app:analysis}

\subsection{The Landscape of Skill Effects}
\label{app:landscape}

The causal attribution matrix, introduced in \S\ref{sec:attribution} as a computational tool, doubles as a diagnostic lens, and what it reveals is, to our knowledge, a finding that no prior work on agent skill libraries has reported.
On the GPT-5.1 attribution (15~development tasks, 103~curated skills), over 90\% of skills have a per-task causal range exceeding $\tau_{\text{split}}{=}0.40$, meaning that nearly every skill in the library helps on some tasks and hurts on others.
Rather than relying solely on this descriptive statistic, we subject the heterogeneity claim to a stricter test: six independent lines of outcome-level evidence, each probing a different consequence that \emph{must} hold if causal heterogeneity is real.
For completeness, Appendix~\ref{app:stat_validation} (Table~\ref{tab:power}) reports that per-cell permutation tests lack adequate power at $M{=}12$, which is precisely why outcome-level validation provides a more rigorous foundation:
\begin{enumerate}[nosep,leftmargin=*]
\item \textbf{Masking diversity.} Per-task masking produces genuinely diverse skill subsets: 96\% of dropped skills are selective (not universally dropped), with mean pairwise Jaccard similarity of only 0.30, ruling out a fixed pruning artefact.
\item \textbf{Attribution alignment.} Dropped skills score significantly worse than kept skills on per-task attribution (gap\,$=$\,0.147, Mann-Whitney $p < 10^{-98}$), confirming that masking decisions track measured causal signal.
\item \textbf{Sign-reversing cases.} Recovered tasks consistently exhibit skills with positive aggregate but negative per-task scores (e.g., \texttt{vc-00109a}: $\bar{C}{=}{+}0.007$ vs.\ $\hat{C}{=}{-}0.107$), the hallmark of causal heterogeneity.
\item \textbf{Split separation.} 7 of 13 splits produce daughter variants with $\geq 0.05$ systematic score divergence, confirming that the split operation captures real conditional structure.
\item \textbf{Reverse masking.} Suppressing the most \emph{positively}-scoring skills instead of the most negatively-scoring ones degrades performance by 4.7\,pp, confirming that the direction of the attribution, not mere context reduction, drives the gain.
\item \textbf{Bootstrap sign stability.} Sign stability averages 71\%/70\% for positive/negative cells among the top-26 heterogeneous skills, well above the 50\% chance baseline.
\end{enumerate}
The remaining $\sim$9\% of skills have global scores near zero; skills with uniformly positive or uniformly negative effects are effectively absent.
The pattern is qualitatively similar across all seven models (library sizes range from 71 to 126).
Figure~\ref{fig:heatmap} visualises this structure directly: nearly every row of the attribution matrix contains both green and red cells.

This finding reframes the curation problem.
The conventional assumption, implicit in all prior work, is that skills are either good or bad, and the curator's job is to separate the two.
Our attribution reveals a different landscape: most skills are \emph{conditionally} good or bad depending on the task, and the real challenge is determining \emph{when} each skill should be active.
This is why per-task masking dominates offline restructuring in the ablation ($+$7.5 vs.\ $+$2.0~pp): the bottleneck is not removing bad skills from the library, but matching skills to tasks at inference time.
Concrete examples of sign-reversing skills are provided in Appendix~\ref{app:skill_examples}.

\subsection{Masking Behaviour}
\label{app:masking_analysis}

The landscape analysis explains \emph{why} per-task masking helps; we now examine \emph{how} it operates in practice.
Across all 417 \textit{test\_challenge} tasks (GPT-5.1), the method suppresses an average of 5.5~skills per task (5.3\% of the library) among the 329~tasks where masking is active, with a standard deviation of 2.8.
On the remaining 88~tasks (21.1\%), the fallback to the full library activates.
Two properties of the suppressed sets are notable.
First, the sets are highly task-dependent: only 2~skills are suppressed on more than 80\% of non-fallback tasks, confirming that masking produces genuinely different libraries for different tasks rather than a fixed pruning.
Second, the suppressed skills are not random: they consistently correspond to skills whose predicted causal scores are strongly negative on the nearest development tasks, confirming that the masking rule selects on measured harm rather than surface features.

A representative example ties the mechanism back to Figure~\ref{fig:masking_example}.
Task \texttt{6474048\_2} (the coffee grinder from \S\ref{sec:intro}) fails both without skills (incorrect dimension filtering) and with the full library (off-topic Spotify and side-effect-checking rules distract the agent).
Per-task masking drops exactly these two irrelevant skills, and the agent succeeds, a microcosm of the broader pattern: measurement identifies the interference, and masking removes it.

\subsection{Limitations}
\label{app:limitations}

Several limitations merit discussion.
GPT-5.1's null result on $\tau$-bench (62.6\%$\to$62.6\%) highlights a boundary of context engineering.
Two factors compound.
First, GPT-5.1 already achieves 100\% on read-only queries without any skill library, yet trails GPT-4.1 by over 10 percentage points on multi-step actions (exchange, modify), suggesting that its internal prior saturates easy tasks but is insufficiently aligned to absorb external procedural knowledge on hard ones.
Second, the skill library is curated on AppWorld, a code-generation environment whose operational patterns transfer only partially to $\tau$-bench's conversational customer-service setting; GPT-5.1, with the strongest internal priors, is the most sensitive to this domain gap.
These two hypotheses make distinguishable predictions.
If the bottleneck is capability saturation, all action types should plateau; Table~\ref{tab:taubench_action} shows otherwise: GPT-5.1 achieves 100\% on read-only but only 56.7\% on exchanges, indicating uneven rather than uniform saturation.
If the bottleneck is domain transfer, the model with the strongest internal priors should be most resistant to external skill injection; GPT-5.1 is indeed the only model showing zero gain, consistent with this prediction.
Claude Sonnet~4.5 independently confirms this pattern on $\tau$-bench (73.0\%$\to$73.0\%).
Its base performance (73.0\%) is comparable to GPT-4.1's augmented result (73.9\%), placing it in the high-competence regime where prompt-time skill injection yields diminishing returns.
Two models from different providers exhibiting the same null result strengthens the interpretation: the boundary is determined by the model's baseline competence on the target domain, not by provider-specific idiosyncrasies.
More broadly, the marginal value of prompt-time skill injection appears to diminish as base model competence strengthens.
The attribution matrix is computed on $M{=}15$ development tasks and extrapolated via nearest-neighbour weighting.
On \textit{test\_normal}, mean cosine similarity to the top-8 nearest development tasks is 0.578, with 21.4\% of tasks below 0.5; on \textit{test\_challenge} these figures are 0.469 and 69.1\% (Appendix~\ref{app:stat_validation}, Figure~\ref{fig:coverage_cdf}).
Despite limited coverage, the method gains $+$4.8\,pp even in the lowest-coverage quartile of \textit{test\_normal}, supported by the structural fallback which activates on 21.1\% of tasks.
The split step invokes LLM judgment to produce conditional skill variants, reintroducing the subjectivity we aim to reduce; we bound this by limiting candidates and validating through the development gate, but a fully measurement-driven splitting procedure remains open.
Finally, our framework assumes a fixed skill library; extending it to online settings where skills are continuously added would require incremental attribution updates.
\section{Operational Templates}
\label{app:templates}

The five AppWorld templates and five $\tau$-bench retail templates below are appended to every skill library as a protected bedrock layer.
They are exempt from all modification and masking operations.
Each template instantiates one of five domain-agnostic operational principles: complete data collection before acting, verify intermediate state before proceeding, confirm before irreversible actions, resolve ambiguous entities from authoritative sources, and sequence dependent operations correctly.
In code-generation agents (AppWorld), templates take the form of executable code scaffolds; in conversational agents ($\tau$-bench), they take the form of procedural checklists derived from the benchmark's policy document.

\paragraph*{AppWorld Templates.}

\noindent\texttt{[tpl-00001] PAGINATION TEMPLATE.}\;
Use for any API that returns a list.
Always paginate with a \texttt{while True} loop over \texttt{page\_index}; never use \texttt{range(N)} or fetch only one page.
Accumulate results into \texttt{all\_items} and break when the page is empty.

\noindent\texttt{[tpl-00002] DATA VALIDATION TEMPLATE.}\;
Before acting on any data, print and inspect intermediate results (item count, sample keys, first item).
Only proceed after confirming the data looks correct.

\noindent\texttt{[tpl-00003] SAFE COMPLETE\_TASK TEMPLATE.}\;
Before submitting an answer, print it alongside the number of source data points from which it was derived.
Only then call \texttt{complete\_task}.

\noindent\texttt{[tpl-00004] LOGIN + ACCESS TOKEN TEMPLATE.}\;
Standard login flow: retrieve account passwords via \texttt{supervisor.show\_account\_passwords}, match by app name, call the app's \texttt{login} endpoint, and store the access token.

\noindent\texttt{[tpl-00005] CROSS-APP IDENTITY RESOLUTION TEMPLATE.}\;
When resolving a person across apps, always start from the phone contacts directory (paginated).
Match by email or phone number, never by name alone.

\paragraph*{$\tau$-bench Retail Templates.}

\noindent\texttt{[tpl-tb-001] ITEM MODIFICATION CHECKLIST.}\;
Before calling \texttt{modify\_pending\_order\_items} or \texttt{exchange\_delivered\_order\_items}: (1)~call \texttt{get\_product\_details} for each item, (2)~present options and confirm exact choice, (3)~ask if these are all the items, (4)~collect all changes into a single list, (5)~make one tool call.
These tools can only be called once.

\noindent\texttt{[tpl-tb-002] EXCHANGE ITEM RESOLUTION.}\;
Six-step procedure: get current item IDs, fetch product details for available variants, match customer requirements, present ambiguous options, compare prices if requested, confirm exact new item ID.

\noindent\texttt{[tpl-tb-003] MULTI-ORDER SEQUENCING.}\;
For requests involving multiple orders: get all order IDs, check each order's status, process sequentially, confirm each action individually, track completed orders.

\noindent\texttt{[tpl-tb-004] STATUS-GATED ACTION.}\;
Before any order action, verify status via \texttt{get\_order\_details}.
Cancel/modify require ``pending''; return/exchange require ``delivered''.
If status does not match, inform the customer of available actions.

\noindent\texttt{[tpl-tb-005] CONFIRMATION PROTOCOL.}\;
Before any destructive action, present a summary of: order ID, specific items and changes, payment method for refund, expected outcome.
Wait for explicit customer confirmation before executing.

\section{Causal Attribution Matrix}
\label{app:heatmap}

\begin{figure}[h]
\centering
\includegraphics[width=0.9\textwidth]{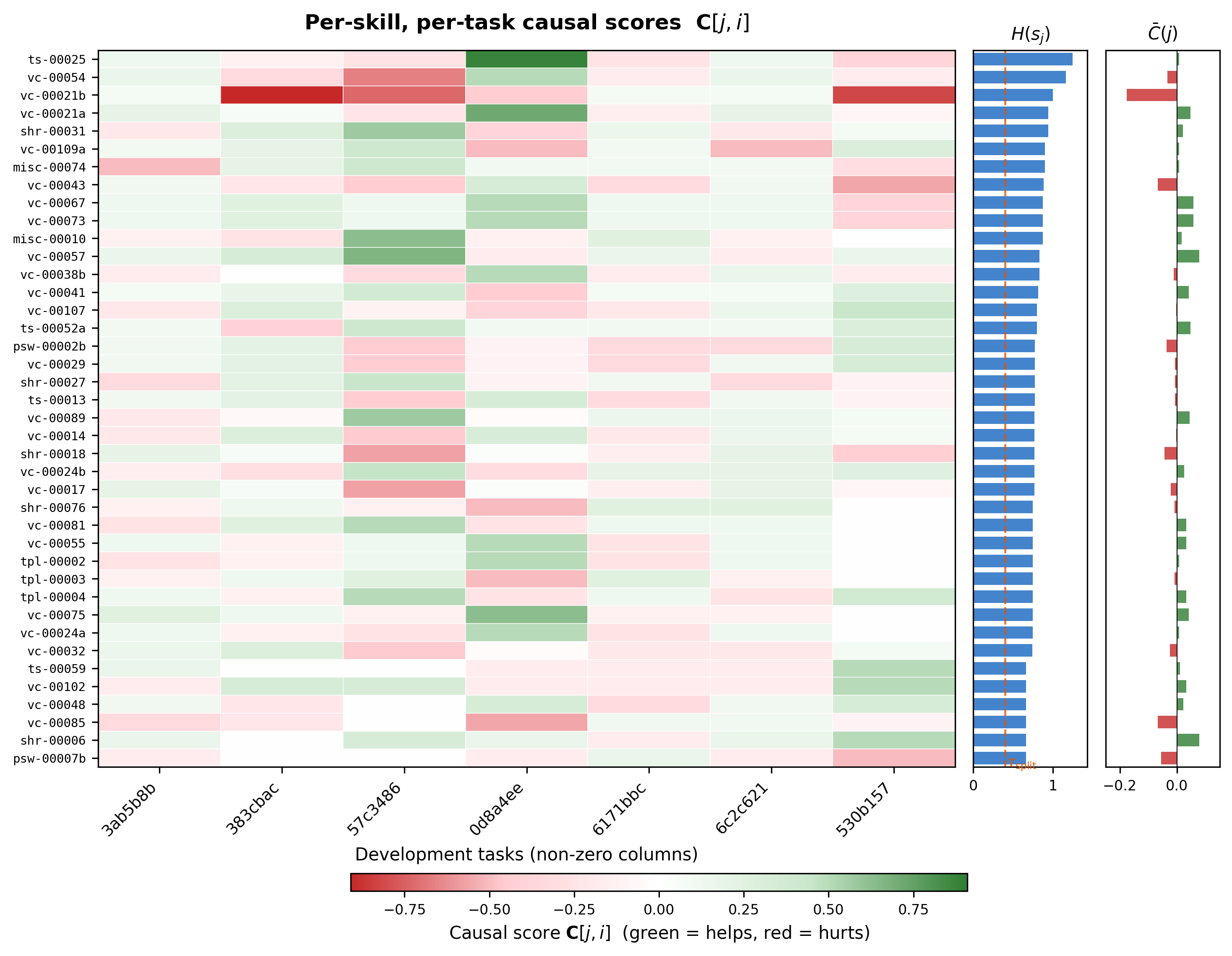}
\caption{\textbf{Causal attribution matrix for GPT-5.1 on AppWorld} (top 40 skills by heterogeneity, 7 informative development tasks).
Each cell shows the causal score $\mathbf{C}[j,i]$: green indicates the skill helps on that task, red indicates harm.
\emph{Left}: the matrix itself, sorted by heterogeneity $H(s_j)$ descending.
Nearly every row contains both positive and negative cells, and outcome-level validation (\S\ref{app:landscape}, Appendix~\ref{app:stat_validation}) confirms this pattern reflects genuine task-dependent effects rather than sampling noise.
\emph{Centre}: the heterogeneity score $H(s_j)$; the dashed orange line marks the split threshold $\tau_{\text{split}}{=}0.40$.
\emph{Right}: the global causal score $\bar{C}(j)$.
Skills with near-zero global scores but wide per-task ranges are the most dangerous, as they are invisible to any curation method that evaluates skills globally.
}
\label{fig:heatmap}
\end{figure}

Figure~\ref{fig:heatmap} visualises the causal attribution matrix for GPT-5.1 on AppWorld, restricted to the 40 most heterogeneous skills and the 7 most informative development tasks.
Each cell encodes the difference-in-means causal score $\mathbf{C}[j,i]$ (Eq.~\ref{eq:causal}): green indicates that including the skill improves task performance, red indicates harm.
The pervasive coexistence of green and red within nearly every row confirms the central empirical finding of this paper (over 90\% of skills are causally heterogeneous, validated through six outcome-level tests in \S\ref{app:landscape} and Appendix~\ref{app:stat_validation}) and motivates per-task masking as the primary curation mechanism.
The right-hand panels display the heterogeneity score $H(s_j)$ and the global causal score $\bar{C}(j)$; skills with high heterogeneity but near-zero global scores are the most dangerous, as they evade any curation method that evaluates skills only in aggregate.

\section{Sign-Reversing Skill Examples}
\label{app:skill_examples}

Two examples from the GPT-5.1 attribution on AppWorld illustrate causal heterogeneity at the individual skill level.

\noindent \textbf{Example 1: \texttt{vc-00054}.}
``Validate that note names map unambiguously to contacts before creating Venmo requests.''
Global causal score: $\bar{C} = -0.03$ (near zero).
Per-task range: $+0.50$ on shared-expense reconciliation tasks (catches real name ambiguities) to $-0.67$ on single-app tasks (forces irrelevant cross-referencing).
A judgment-based curator would retain this skill because its global score is harmless; only per-task measurement reveals that it helps half the time and hurts the other half.

\noindent \textbf{Example 2: \texttt{vc-00021a}.}
``For repeated side-effecting API calls, capture and check each response.''
Global causal score: $\bar{C} = +0.05$ (near zero).
Per-task range: $+0.71$ on mutation-heavy tasks (catches silent API failures) to $-0.23$ on read-only tasks (adds pure overhead).
Both examples share the signature of causal heterogeneity: near-zero global score, large per-task range, and invisible to any curation method that does not disaggregate by task.

\section{Per-Difficulty Breakdown}
\label{app:perlevel}

Table~\ref{tab:perlevel} disaggregates Task Goal Completion by difficulty level (L1--L3) for six base models on both AppWorld test splits.
Level~1 tasks are largely saturated across methods, while the most substantial improvements from our pipeline appear on Level~2 and Level~3 tasks---precisely the multi-step scenarios where uncurated skill libraries cause the most interference.
For GPT-5.1 on \textit{test\_challenge}, the upstream ACE library degrades L2 and L3 performance relative to bare ReAct (50.0$\to$44.0 and 48.2$\to$43.1), while our method recovers and substantially exceeds baseline performance at every difficulty level.

\begin{table}[H]
\caption{\textbf{Per-difficulty breakdown} (TGC).
Gains concentrate on Level~2 and Level~3 tasks.
GPT-4o per-level breakdown is not available at single-attempt granularity.}
\label{tab:perlevel}
\centering
\small
\setlength{\tabcolsep}{4pt}
\renewcommand{\arraystretch}{1.05}
\begin{tabular}{@{}ll cccc cccc@{}}
\toprule
& & \multicolumn{4}{c}{\textit{test\_normal}} & \multicolumn{4}{c}{\textit{test\_challenge}} \\
\cmidrule(lr){3-6} \cmidrule(lr){7-10}
\textbf{Model} & \textbf{Method} & L1 & L2 & L3 & All & L1 & L2 & L3 & All \\
\midrule
\multirow{3}{*}{\small GPT-5.1}
 & ReAct         & 86.0 & 68.8 & 44.4 & 61.9  & 69.4 & 50.0 & 48.2 & 52.5 \\
 & ACE           & 86.0 & 72.9 & 46.0 & 67.3  & 80.6 & 44.0 & 43.1 & 49.9 \\
 & Ours          & \textbf{93.0} & \textbf{93.8} & \textbf{60.3} & \textbf{77.4}
                 & \textbf{88.9} & \textbf{68.7} & \textbf{71.3} & \textbf{66.4} \\
\addlinespace[3pt]
\multirow{3}{*}{\small DS-V3}
 & ReAct            & 89.5 & 75.0 & 46.0 & 69.1  & 61.1 & 45.3 & 43.1 & 47.0 \\
 & ACE              & 93.0 & 85.4 & 58.7 & 78.0  & 77.8 & 61.3 & 59.0 & 63.1 \\
 & Ours             & 93.0 & \textbf{91.7} & \textbf{68.3} & \textbf{83.3}
                    & \textbf{86.1} & \textbf{72.7} & 60.5 & \textbf{69.3} \\
\addlinespace[3pt]
\multirow{3}{*}{\small GPT-4.1}
 & ReAct         & 84.2 & 79.2 & 41.3 & 66.7  & 79.2 & 43.3 & 45.1 & 50.4 \\
 & CUGA          & 91.2 & 77.1 & 54.0 & 73.2  & 91.7 & 58.7 & 44.1 & 57.6 \\
 & Ours          & 91.2 & \textbf{85.4} & 54.0 & \textbf{75.6}
                 & 75.0 & \textbf{61.3} & \textbf{62.1} & \textbf{64.0} \\
\addlinespace[3pt]
\multirow{2}{*}{\small GPT-5.4}
 & ReAct         & 85.7 & 82.1 & 78.6 & 82.1  & 83.5 & 81.3 & 78.4 & 81.1 \\
 & Ours          & \textbf{91.1} & \textbf{87.5} & \textbf{87.5} & \textbf{88.7}
                 & \textbf{87.1} & \textbf{84.2} & \textbf{84.9} & \textbf{85.4} \\
\addlinespace[3pt]
\multirow{2}{*}{\small Sonnet 4.5}
 & ReAct         & 87.5 & 80.4 & 83.9 & 83.9  & 71.2 & 71.9 & 67.6 & 70.3 \\
 & Ours          & \textbf{94.6} & \textbf{85.7} & \textbf{87.5} & \textbf{89.3}
                 & \textbf{75.5} & \textbf{76.3} & \textbf{74.1} & \textbf{75.3} \\
\addlinespace[3pt]
\multirow{2}{*}{\small Gemini 2.5}
 & ReAct         & 76.8 & 69.6 & 71.4 & 72.6  & 48.2 & 46.8 & 53.2 & 49.4 \\
 & Ours          & \textbf{80.4} & \textbf{80.4} & \textbf{82.1} & \textbf{81.0}
                 & \textbf{56.1} & \textbf{54.0} & \textbf{54.7} & \textbf{54.9} \\
\bottomrule
\end{tabular}
\end{table}

\section{Per-Action Breakdown ($\tau$-bench)}
\label{app:peraction}

Table~\ref{tab:taubench_action} reports per-action accuracy on $\tau$-bench retail for all seven models under our method.
The five action categories---exchange, return, modify, cancel, and read-only---differ substantially in procedural complexity: read-only queries require no state mutation, while exchanges involve multi-step item resolution, price comparison, and confirmation.
All models achieve near-perfect accuracy on read-only queries (except GPT-4o), and the largest gains from skill-based curation concentrate on the more complex action types (return and cancel), where procedural templates and curated skills provide the strongest marginal value over unaugmented baselines.

\begin{table}[H]
\caption{\textbf{$\tau$-bench per-action breakdown} (our method). Task counts in parentheses. Models sorted by overall score.}
\label{tab:taubench_action}
\centering
\small
\setlength{\tabcolsep}{6pt}
\renewcommand{\arraystretch}{1.15}
\begin{tabular}{@{}l ccccc c@{}}
\toprule
\textbf{Model} & \makecell{\textbf{Exchange}\\(30)} & \makecell{\textbf{Return}\\(31)} & \makecell{\textbf{Modify}\\(34)} & \makecell{\textbf{Cancel}\\(11)} & \makecell{\textbf{Read-}\\[-2pt]\textbf{only} (9)} & \textbf{All} \\
\midrule
\rowcolor{oursHL}
GPT-5.4    & 73.3 & 80.6 & 79.4 & \textbf{87.5} & \textbf{100}  & \textbf{80.9} \\
GPT-4.1    & 70.0 & \textbf{83.9} & 67.6 & 81.8 & \textbf{100}  & 73.9 \\
Gemini 2.5 & \textbf{76.7} & 71.0 & 64.7 & \textbf{87.5} & 91.7 & 73.9 \\
Sonnet 4.5 & 66.7 & 71.0 & 70.6 & \textbf{87.5} & 91.7 & 73.0 \\
DS-V3      & 70.0 & 64.5 & 61.8 & 63.6 & \textbf{100}  & 66.1 \\
GPT-5.1    & 56.7 & 64.5 & 55.9 & 63.6 & \textbf{100}  & 62.6 \\
GPT-4o     & 66.7 & 77.4 & 61.8 & 27.3 & 44.4 & 62.6 \\
\bottomrule
\end{tabular}
\end{table}

\section{Reproducibility Details}
\label{app:reproducibility}

This section provides the complete set of hyperparameters, model identifiers, and software versions needed to reproduce all experiments.
Table~\ref{tab:hyperparams} lists every hyperparameter used across the pipeline; all values are fixed across every (model, benchmark) configuration with no per-cell tuning.
Table~\ref{tab:api} lists the exact API identifiers for all models.

\begin{table}[H]
\caption{\textbf{Hyperparameters.} All values are fixed across every (model, benchmark) cell.}
\label{tab:hyperparams}
\centering
\small
\setlength{\tabcolsep}{8pt}
\renewcommand{\arraystretch}{1.15}
\begin{tabular}{@{}l l c@{}}
\toprule
\rowcolor{headerBG}
\textbf{Stage} & \textbf{Parameter} & \textbf{Value} \\
\midrule
\multirow{2}{*}{Upstream curation}
 & Difficulty-estimation rollouts & 2 \\
 & Task ordering & Hardest-first \\
\addlinespace[2pt]
\rowcolor{stripeBG}
\multirow{3}{*}{Causal attribution}
 & \cellcolor{stripeBG} Development tasks $M$ & \cellcolor{stripeBG} 15 \\
 \rowcolor{stripeBG}
 & Random masks $K$ & 12 \\
 \rowcolor{stripeBG}
 & Keep probability $f$ & 0.4 \\
\addlinespace[2pt]
\multirow{4}{*}{Offline restructuring}
 & Split threshold $\tau_{\text{split}}$ & 0.40 \\
 & Max split candidates & 15 \\
 & Retire threshold $\tau_{\text{retire}}$ & 0.10 \\
 & Merge similarity $\tau_{\text{merge}}$ & 0.85 \\
\addlinespace[2pt]
\rowcolor{stripeBG}
\multirow{4}{*}{Per-task masking}
 & \cellcolor{stripeBG} Nearest neighbours $k$ & \cellcolor{stripeBG} 8 \\
 \rowcolor{stripeBG}
 & Softmax temperature $\tau$ & 5 \\
 \rowcolor{stripeBG}
 & Mask threshold $\tau_{\text{mask}}$ & $-$0.10 \\
 \rowcolor{stripeBG}
 & Min retained skills & 30 \\
\addlinespace[2pt]
Shared & Embedding model & Qwen3-Emb-0.6B \\
\bottomrule
\end{tabular}
\end{table}

\noindent \textbf{Model identifiers.}
All models are accessed via API; Table~\ref{tab:api} lists the exact identifiers returned by each provider.
\begin{table}[H]
\caption{\textbf{Model API identifiers.}}
\label{tab:api}
\centering
\small
\begin{tabular}{@{}l l l@{}}
\toprule
\textbf{Model} & \textbf{API identifier} & \textbf{Provider} \\
\midrule
GPT-5.1     & \texttt{gpt-5.1-2025-11-13}  & Azure OpenAI \\
GPT-5.4     & \texttt{gpt-5.4-2026-03-05}  & Azure OpenAI \\
GPT-4.1     & \texttt{gpt-4.1-2025-04-14}  & Azure OpenAI \\
GPT-4o      & \texttt{gpt-4o-2024-11-20}\textsuperscript{$\dagger$}   & Azure OpenAI \\
DeepSeek-V3 & \texttt{deepseek-chat} (V3.2) & DeepSeek direct \\
Sonnet 4.5  & \texttt{claude-sonnet-4-5-20250929} & Anthropic direct \\
Gemini 2.5 Pro & \texttt{gemini-2.5-pro}   & Google AI Studio \\
\midrule
Embedding   & Qwen3-Embedding-0.6B         & HuggingFace (local) \\
$\tau$-bench user sim & \texttt{gpt-4o-2024-11-20} & Azure OpenAI \\
\bottomrule
\end{tabular}
\vspace{2pt}
\par\noindent{\footnotesize $^\dagger$Our agent and the \citet{gupta2025leveraging} baseline use this version.
The ReAct baseline (48.8\%) is from the AppWorld leaderboard entry using \texttt{gpt-4o-2024-05-13}.}
\end{table}

\noindent \textbf{Benchmark versions.}
AppWorld: \texttt{appworld==0.1.4.dev0} (pip, aligned with the ACE codebase).
$\tau$-bench: commit \texttt{4754e6b} of \texttt{sierra-research/tau-bench}.

\noindent \textbf{Determinism.}
All agent calls use temperature~0 and seed~100.
Attribution masks are generated with Python's \texttt{random.Random(seed)} where seed$\,{=}\,$42 for all (model, benchmark) cells except GPT-5.1 on AppWorld, which uses seed$\,{=}\,$45.
Attribution configuration is otherwise uniform: $K{=}12$ masks, $M{=}15$ development tasks, keep probability $f{=}0.4$ (Bernoulli).

\noindent \textbf{Library sizes.}
After upstream curation (before offline restructuring), the skill library contains 103~skills for GPT-5.1, 87 for GPT-4.1, 126 for GPT-4o, 80 for DeepSeek-V3, 95 for GPT-5.4, 71 for Claude Sonnet~4.5, and 88 for Gemini 2.5 Pro on AppWorld.
On $\tau$-bench, library sizes are 76, 72, 61, 161, 81, 94, and 68 respectively.
DeepSeek-V3's notably larger $\tau$-bench library (161 vs.\ 61--76 for the OpenAI models) reflects more aggressive skill generation during curation: DeepSeek-V3 proposed 115 ADD actions over 500 training tasks, compared to 15--30 for the other models under the same curation prompt.
Our downstream attribution and masking mechanisms are agnostic to library size.
Variation arises because each model's curation run produces a different set of skills from the same training tasks.
\section{Sequential Ablation}
\label{app:ablation}

Table~\ref{tab:ablation} isolates the contribution of each pipeline component by sequentially adding them to the bare ReAct baseline on GPT-5.1 / AppWorld \textit{test\_normal} (168 tasks).
This additive design ensures that each row's $\Delta$ reflects the marginal value of the newly added component on top of all preceding ones.
The results confirm that per-task masking contributes the largest single increment ($+$7.5\,pp), consistent with the finding that the dominant challenge is not removing globally harmful skills but selecting the right skill subset for each individual task.

\begin{table}[H]
\caption{\textbf{Sequential ablation} (GPT-5.1, AppWorld \textit{test\_normal}, 168 tasks).
Each row adds one component to the preceding configuration.
Per-task masking contributes the largest single increment ($+$7.5\,pp), consistent with the finding that over 90\% of skills are causally heterogeneous (\S\ref{app:landscape}).}
\label{tab:ablation}
\centering
\small
\setlength{\tabcolsep}{10pt}
\renewcommand{\arraystretch}{1.15}
\begin{tabular}{@{}l cc@{}}
\toprule
\textbf{Configuration} & \textbf{TGC} & $\boldsymbol{\Delta}$ \\
\midrule
(\textit{a})\; ReAct baseline             & 61.9 & --- \\
(\textit{b})\; + Templates                & 67.9 & {\color{posdelta}$+$6.0} \\
(\textit{c})\; + Offline Restructuring    & 69.9 & {\color{posdelta}$+$2.0} \\
\rowcolor{oursHL}
(\textit{d})\; + Per-Task Masking (full)  & \textbf{77.4} & {\color{posdelta}$+$7.5} \\
\bottomrule
\end{tabular}
\end{table}
\section{Offline Restructuring: Split Example}
\label{app:split_example}

To illustrate what offline restructuring produces in practice, we show one of the 13 splits performed on the GPT-5.1 AppWorld library (bifurcation score $H = 1.56$).

\noindent \textbf{Before (single skill).}
\texttt{[psw-00007]} ``Many APIs return items in pages. Make sure to run through all the pages by looping over \texttt{page\_index}.''

This rule is essential for tasks that require scanning many items (e.g., browsing products, enumerating emails), but harmful for targeted purchases where the agent already knows which item it wants: exhaustive pagination wastes the step budget.

\noindent \textbf{After (two conditional variants).}
\texttt{[psw-00007a]} \textit{[IF task involves browsing or comparing multiple product options or scanning through multiple emails/messages]:} ``Many APIs return items in pages. Make sure to run through all the pages by looping over \texttt{page\_index}.''

\texttt{[psw-00007b]} \textit{[IF task is a targeted purchase of a specific known item or constrained to prior sellers]:} ``Do NOT run through all pages or loop over \texttt{page\_index}; instead, focus only on the specific item or on items from previously used sellers without exhaustive pagination.''

Both variants pass the development gate (no regression on any of the $M{=}15$ attribution tasks).
The trigger conditions are generated by the base LLM from the per-task causal score vector; the development gate ensures they do not introduce regressions.

\section{Sub-Goal Completion (SGC)}
\label{app:sgc}

Table~\ref{tab:appworld_sgc} reports Sub-Goal Completion on AppWorld, a finer-grained metric that awards partial credit for completing individual sub-goals within each task even when the overall task fails.
SGC complements the primary TGC metric by revealing whether improvements reflect more tasks being fully solved or deeper progress on partially solved ones.
The pattern mirrors the TGC results: our method achieves the highest SGC for six of seven models on both splits, with the largest gains on \textit{test\_challenge} where uncurated libraries cause the most interference.

\begin{table}[h]
\caption{\textbf{AppWorld SGC results} (\%).
\textbf{Bold}: best prompt-based method per column.
{\color{negdelta}Red}: degrades over ReAct.}
\label{tab:appworld_sgc}
\centering
\small
\setlength{\tabcolsep}{7pt}
\renewcommand{\arraystretch}{1.15}
\begin{tabular}{@{}ll cc@{}}
\toprule
& & \textit{test\_normal} (168) & \textit{test\_challenge} (417) \\
\cmidrule(lr){3-3} \cmidrule(lr){4-4}
\textbf{Model} & \textbf{Method} & SGC & SGC \\
\midrule
\multirow{3}{*}{GPT-5.1}
 & ReAct               & 46.4 & 30.9 \\
 & ACE                 & 55.4 & {\color{negdelta}28.8} \\
 \rowcolor{oursHL}
 & \textbf{Ours}       & \textbf{71.4} & \textbf{48.9} \\
\addlinespace[3pt]
\multirow{3}{*}{DeepSeek-V3}
 & ReAct               & 48.2 & 25.2 \\
 & ACE                 & 66.1 & 43.9 \\
 \rowcolor{oursHL}
 & \textbf{Ours}       & \textbf{71.4} & \textbf{52.5} \\
\addlinespace[3pt]
\multirow{3}{*}{GPT-4.1}
 & ReAct               & 46.4 & 32.4 \\
 & CUGA                & \textbf{62.5} & \textbf{48.2} \\
 \rowcolor{oursHL}
 & \textbf{Ours}       & 60.7 & 48.2 \\
\addlinespace[3pt]
\multirow{3}{*}{GPT-4o}
 & ReAct               & 32.1 & 13.0 \\
 & \citeauthor{gupta2025leveraging} & 57.1 & 23.0 \\
 \rowcolor{oursHL}
 & \textbf{Ours}       & \textbf{58.9} & \textbf{30.2} \\
\addlinespace[3pt]
GPT-5.4
 & ReAct               & 91.9 & 91.7 \\
 \rowcolor{oursHL}
 & \textbf{Ours}       & \textbf{96.7} & \textbf{95.4} \\
\addlinespace[3pt]
Sonnet 4.5
 & ReAct               & 94.0 & 88.7 \\
 \rowcolor{oursHL}
 & \textbf{Ours}       & \textbf{97.2} & \textbf{93.5} \\
\addlinespace[3pt]
Gemini 2.5
 & ReAct               & 87.8 & 80.5 \\
 \rowcolor{oursHL}
 & \textbf{Ours}       & \textbf{95.5} & \textbf{91.8} \\
\midrule
\rowcolor{leaderHL}
\multicolumn{2}{@{}l}{\small\textit{Leaderboard best$^*$ (Qwen3-14B, wt-tuned)}}
 & \textit{80.4} & \textit{50.4} \\
\bottomrule
\multicolumn{4}{@{}l}{\footnotesize $^*$No accompanying publication; see footnote in \S\ref{sec:setup}.}
\end{tabular}
\end{table}

\section{Statistical Validation}
\label{app:stat_validation}

Statistical validation is conducted on GPT-5.1; the same attribution protocol ($K{=}12$, $M{=}15$, $f{=}0.4$) applies to all seven models.

\subsection{Power Analysis}
\label{app:power}

Table~\ref{tab:power} reports the statistical power of a per-skill permutation test for detecting a true $\pm 0.30$ causal effect at the $\alpha{=}0.05$ level, as a function of the number of random masks $M$.
At $M{=}12$ (our setting), per-cell permutation tests achieve only 38.5\% power, and the descriptive threshold $H \geq 0.40$ cannot distinguish real heterogeneity from noise at the single-skill level.
This is precisely why we adopt a higher evidentiary standard: rather than relying on per-cell significance, we validate heterogeneity through six independent outcome-level tests (\S\ref{app:landscape}) that are robust to the per-cell noise level.
The table also shows that increasing $M$ to 30--50 masks would bring per-cell power to 80--98\%, providing a clear path for future work to strengthen the per-skill analysis.

\begin{table}[H]
\caption{\textbf{Power analysis for per-skill heterogeneity detection.}
Power is computed for a true $\pm 0.30$ effect via 1000-trial permutation simulation.}
\label{tab:power}
\centering
\small
\setlength{\tabcolsep}{8pt}
\renewcommand{\arraystretch}{1.10}
\begin{tabular}{@{}cccc@{}}
\toprule
\textbf{Masks} $M$ & \textbf{Per-cell} $\sigma$ & \textbf{Power ($\pm$0.30)} & \textbf{Null $H{\geq}0.40$ FPR} \\
\midrule
12 (ours) & 0.290 & 38.5\% & $\sim$100\% \\
20        & 0.225 & 59.5\% & --- \\
30        & 0.183 & 80.0\% & --- \\
50        & 0.142 & 98.0\% & --- \\
\bottomrule
\end{tabular}
\end{table}

\subsection{Bootstrap Decision Stability}
\label{app:bootstrap}

To assess the robustness of individual curation decisions, we resample the $K{=}12$ attribution masks with replacement 1000 times and recompute each decision.
Table~\ref{tab:bootstrap} summarises the results.
Among the 287 (skill, task) cells where per-task masking suppresses a skill ($\hat{C} < -0.10$), 97.2\% remain below the threshold at the bootstrap median, indicating strong directional stability---the method consistently identifies the same skills as harmful for each task.
The $k{=}8$ averaging in Eq.~\ref{eq:predicted} is key to this stability: it reduces the effective per-decision noise from $\sigma \approx 0.29$ (per cell) to $\sigma \approx 0.10$ (per decision), bringing the signal-to-noise ratio close to~1 for the masking threshold.
At the stricter 95\% CI level, 4.2\% of cells are confirmed, reflecting the inherent conservatism of cell-level confidence intervals at $M{=}12$; the directional stability rate is the operationally relevant metric since the fallback mechanism (\S\ref{sec:masking}) ensures graceful degradation for borderline cases.

\begin{table}[H]
\caption{\textbf{Bootstrap decision stability} (1000 resamples).}
\label{tab:bootstrap}
\centering
\small
\setlength{\tabcolsep}{6pt}
\renewcommand{\arraystretch}{1.10}
\begin{tabular}{@{}lccc@{}}
\toprule
\textbf{Decision type} & \textbf{Total} & \textbf{Directionally stable} & \textbf{Strictly stable} \\
& & (median) & (95\% CI) \\
\midrule
Mask ($\hat{C} < -0.10$) & 287 cells & 97.2\% & 4.2\% \\
Retire ($|\bar{C}| < 0.10$) & 100 skills & 88.0\% ($\geq$80\% boots) & --- \\
\bottomrule
\end{tabular}
\end{table}

\subsection{Outcome-Level Heterogeneity Evidence}
\label{app:outcome_evidence}

Since per-cell significance testing lacks power at $M{=}12$ (\S\ref{app:power}), we validate the heterogeneity claim through six independent lines of outcome-level evidence, summarised in Table~\ref{tab:outcome_evidence}.
Each line tests a different prediction of the causal heterogeneity hypothesis; together they provide convergent support that the attribution matrix captures genuine task-dependent structure.

\begin{table}[H]
\caption{\textbf{Outcome-level evidence for causal heterogeneity.}}
\label{tab:outcome_evidence}
\centering
\small
\setlength{\tabcolsep}{5pt}
\renewcommand{\arraystretch}{1.10}
\begin{tabular}{@{}l l@{}}
\toprule
\textbf{Evidence} & \textbf{Statistic} \\
\midrule
Mask diversity & 96\% selective, mean Jaccard\,$=$\,0.30 \\
Drop vs.\ Keep attribution & gap\,$=$\,0.147, Mann-Whitney $p < 10^{-98}$ \\
Sign-reversing cases & \texttt{vc-00109a}: $\bar{C}{=}{+}0.007$, $\hat{C}{=}{-}0.107$ \\
Split separation & 7/13 splits $\geq 0.05$ divergence \\
Reverse masking & $-$4.7\,pp (causal direction confirmed) \\
Sign stability & 71\%/70\% pos/neg (vs.\ 50\% null) \\
\bottomrule
\end{tabular}
\end{table}

\subsection{Development Set Coverage}
\label{app:coverage}

\begin{figure}[H]
\centering
\includegraphics[width=0.7\textwidth]{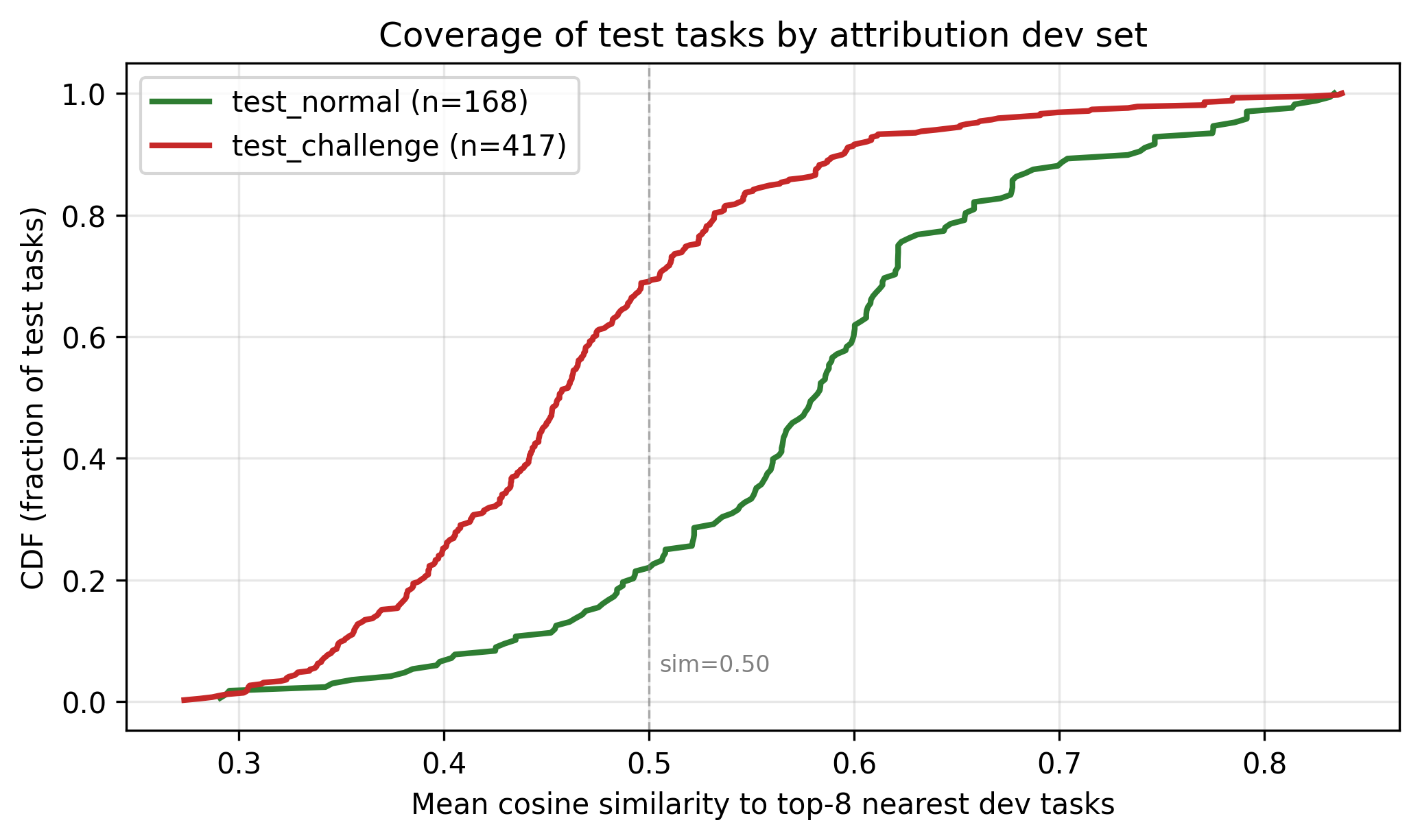}
\caption{\textbf{Coverage of test tasks by the attribution development set.}
CDF of mean cosine similarity to the top-8 nearest development tasks.
The dashed line marks similarity\,$=$\,0.50.}
\label{fig:coverage_cdf}
\end{figure}

Figure~\ref{fig:coverage_cdf} shows the cumulative distribution of mean cosine similarity between each test task and its $k{=}8$ nearest development tasks.
On \textit{test\_normal}, 78.6\% of tasks have similarity $\geq 0.50$; on \textit{test\_challenge}, only 30.9\% do.
Table~\ref{tab:coverage_stratified} stratifies performance by coverage quartile on \textit{test\_normal} (GPT-5.1).
The method improves over ReAct even in the lowest-coverage quartile ($+$4.8\,pp), though gains are largest in the mid-range where the attribution signal is strongest.

\begin{table}[H]
\caption{\textbf{Performance by coverage quartile} (GPT-5.1, \textit{test\_normal}, 168 tasks).}
\label{tab:coverage_stratified}
\centering
\small
\setlength{\tabcolsep}{6pt}
\renewcommand{\arraystretch}{1.10}
\begin{tabular}{@{}l c cc c@{}}
\toprule
\textbf{Quartile} & \textbf{Sim range} & \textbf{ReAct} & \textbf{Ours} & $\boldsymbol{\Delta}$ \\
\midrule
Q1 (low)  & 0.29--0.51 & 61.9\% & 66.7\% & {\color{posdelta}$+$4.8} \\
Q2        & 0.52--0.58 & 59.5\% & 78.6\% & {\color{posdelta}$+$19.0} \\
Q3        & 0.58--0.62 & 69.0\% & 85.7\% & {\color{posdelta}$+$16.7} \\
Q4 (high) & 0.62--0.83 & 71.4\% & 78.6\% & {\color{posdelta}$+$7.1} \\
\bottomrule
\end{tabular}
\end{table}

\end{document}